\documentclass[journal]{IEEEtran}

\usepackage{cite}
\usepackage{graphicx}
\usepackage{float}
\usepackage{caption}
\usepackage{algorithmic}
\usepackage{amsmath,amssymb,amsfonts}
\usepackage{xcolor,float}
\usepackage{subfigure}
\usepackage{verbatim}
\usepackage{type1cm}
\usepackage{soul}
\usepackage{bm}

\IEEEoverridecommandlockouts                              

\begin{document}

\title{
	A 3-Step Optimization Framework with Hybrid Models for a Humanoid Robot's Jump Motion
}

\author{Haoxiang~Qi, Zhangguo~Yu, Xuechao~Chen, Yaliang~Liu, Chuanku~Yi, Chencheng~Dong, \\ Fei~Meng, and Qiang Huang, \emph{Fellow, IEEE}
	\thanks{*This work has been submitted to the IEEE for possible publication. Copyright may be transferred without notice, after which this version may no longer be accessible.}
	\thanks{*This work was supported by the National Natural Science Foundation of China under Grant 91748202 and National Natural Science Foundation of China under Grant 62073041.}
	\thanks{H. Qi, Y. Liu, C. Yi and C. Dong are with the Intelligent Robotics Institute, School of Mechatronical Engineering, Beijing Institute of Technology, Beijing 100081, China (e-mail: 3120215098@bit.edu.cn; liuyaliang@bit.edu.cn; 3120210155@bit.edu.cn;  3120195094@bit.edu.cn).}
	\thanks{Z. Yu, X. Chen, F. Meng and Q. Huang are with the Intelligent Robotics Institute, School of Mechatronical Engineering, Beijing Institute of Technology, Beijing 100081, China, the International Joint Research Laboratory of Biomimetic Robots and Systems, Ministry of Education, Beijing 100081, China, and also with the National Key Lab of Autonomous Intelligent Unmanned Systems, Beijing 100081, China (e-mail: chenxuechao@bit.edu.cn; yuzg@bit.edu.cn; mfly0208@bit.edu.cn; qhuang@bit.edu.cn).}
}

\maketitle
\thispagestyle{empty}
\pagestyle{empty}

\begin{abstract}
	High dynamic jump motions are challenging tasks for humanoid robots to achieve environment adaptation and obstacle crossing. The trajectory optimization is a practical method to achieve high-dynamic and explosive jumping. This paper proposes a 3-step trajectory optimization framework for generating a jump motion for a humanoid robot. To improve iteration speed and achieve ideal performance, the framework comprises three sub-optimizations. The first optimization incorporates momentum, inertia, and center of pressure (CoP), treating the robot as a static reaction momentum pendulum (SRMP) model to generate corresponding trajectories. The second optimization maps these trajectories to joint space using effective Quadratic Programming (QP) solvers. Finally, the third optimization generates whole-body joint trajectories utilizing trajectories generated by previous parts. With the combined consideration of momentum and inertia, the robot achieves agile forward jump motions. A simulation and experiments (Fig. \ref{Fig First page fig}) of forward jump with a distance of 1.0 m and 0.5 m height are presented in this paper, validating the applicability of the proposed framework.
\end{abstract} 

\def\abstractname{Note to Practitioners}
\begin{abstract}
	The motivation of this paper stems from the need to improve jumping performance of humanoid robots. By comprehensively considering factors such as robot posture, centroidal angular momentum, and landing foot placement, the algorithm enhances the robot's ability to navigate complex environments. This capability is crucial for applications that require overcoming obstacles, such as in search and rescue or inspection tasks. Improved jumping ability can significantly boost environmental adaptability, allowing robots to perform effectively in diverse conditions, and it also represents an exploration of the high-dynamic motion capabilities of humanoid robots. Future research will focus on integrating visual and perceptual information to enhance decision-making.
\end{abstract}


\section{Introduction}
\label{Sec Introduction}
\IEEEPARstart{N}{owadays}, researches on humanoid robots has been a significant branch in the field of robotics. With the expanding of the application scenarios for humanoid robots, there is a growing demand for robots' agile mobility. In recent years, researchers have been putting efforts on achieving high-dynamic movements on humanoid robots, such as running, jumping, and acrobatics. For locomotion tasks in unstructured environments, jumping is significant for robots to overcome low obstacles or wide ditches. Additionally, jumping serves as an ideal way to demonstrate a robot's dynamic performance. Atlas from Boston Dynamics has achieved agile jump motions\cite{AtlasWeb}, yet no supplemental article about detailed algorithm was provided. In our previous work \cite{qi2023vertical,haoxiang2020vertical}, vertical jump is achieved on an adult-sized humanoid robot platform. In this paper, we continue our exploration of agile jumping motion for humanoid robot.

\begin{figure}
	\centering
	\includegraphics[scale=0.4]{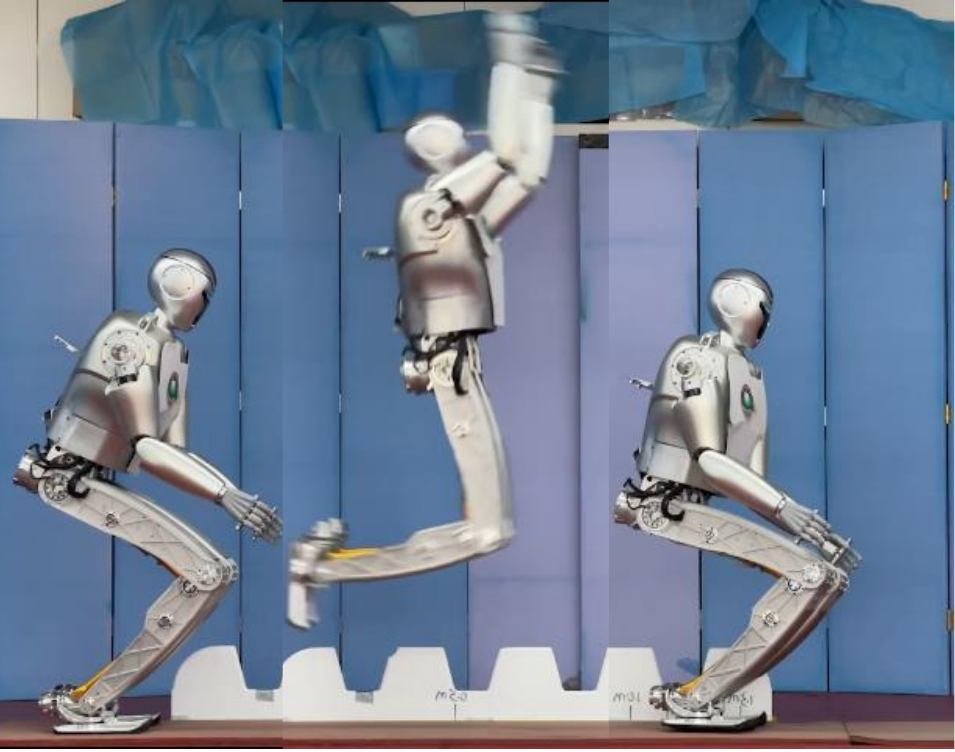}
	\caption{Illustration of a forward jump experiment.}
	\label{Fig First page fig}
\end{figure}

\begin{figure*}
	\centering
	\includegraphics[scale=0.32]{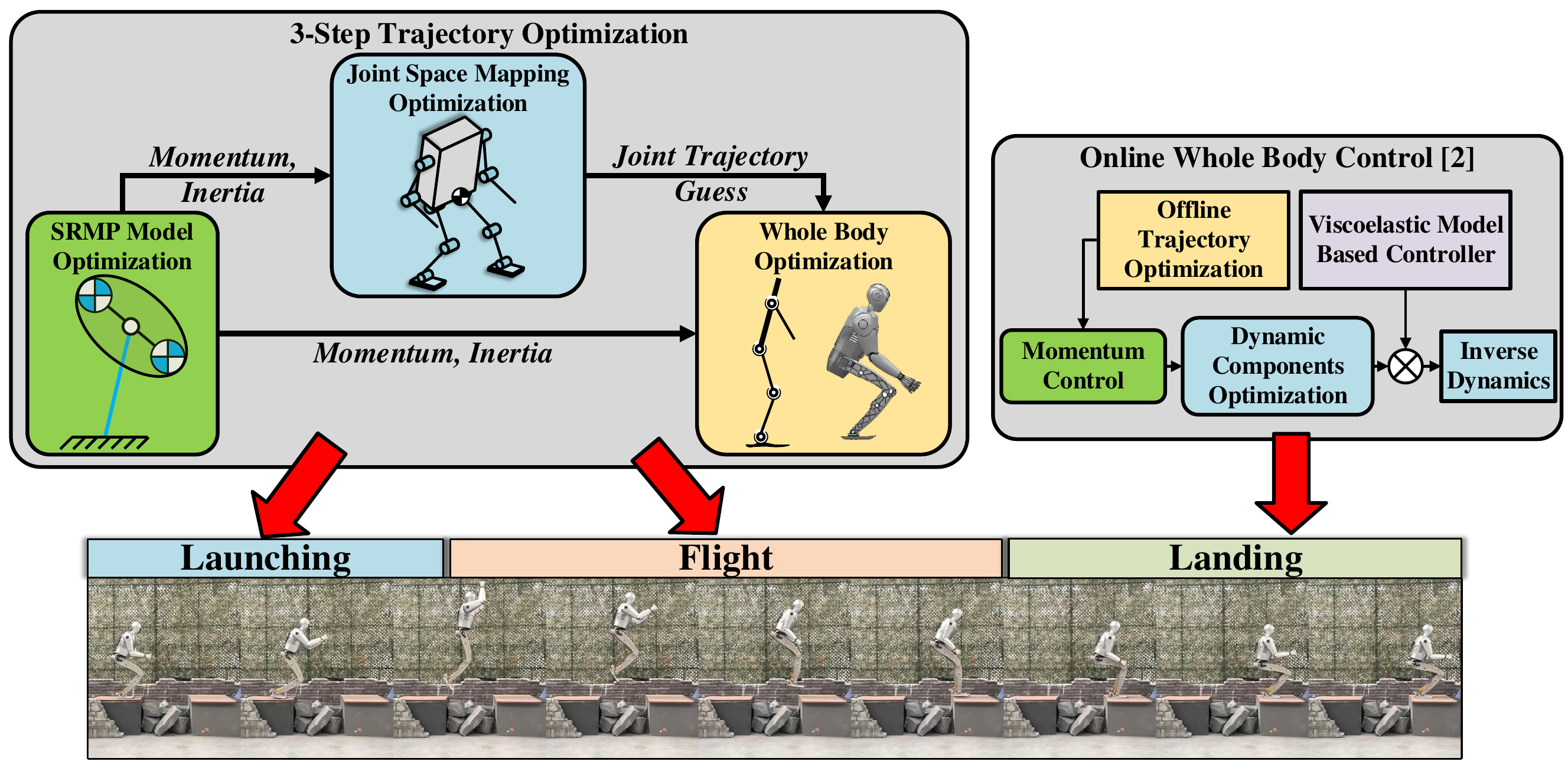}
	\caption{Overview of the framework in this paper}
	\label{Fig Overview of the framework}
\end{figure*}

During the launching phase of a jump motion, the robot must attain a pre-designed posture and centroidal angular momentum (CAM) at the takeoff moment to generate the desired rotation in the air. However, the variation of the posture and CAM are coupled in robot control. Non-zero CAM will cause the robot to rotate around its center of mass (CoM), leading to changes in its body posture. In the field of biped jump research, the inconsistency between the posture and the CAM control has rarely been discussed. In some early researches, keeping the balance and improving the explosive output were the core contributions\cite{ohnishi2004motion,nunez2005control,sripada2018biped,janardhan2013kinematic}. Then, some researchers began to work on the control of the CAM to achieve better performance of the jump motion\cite{wensing2014development,bergonti2019torque,xiong2018bipedal,wensing2016improved,xiong2020sequential,nguyen2019optimized,chignoli2021humanoid}. Simplified model based control, whole body control and control strategies incorporating with the robot's hardware design were explored, and yet the harmonious control of the posture and the CAM is still worth digging. On the other hand, large-scale nonlinear optimization solvers have been widely utilized in the trajectory generation of humanoid robot. Researchers constracted the solver with whole body model and designed complex cost function and constraints to obtain ideal trajectories\cite{xiong2020sequential,dai2014whole,mesesan2023unified,meng2023online}. Yet the multiphase and hybrid contact situation of jump motion can lead to heavy computational burden for such method. In recent years, strategies based on machine learning have shown their advantages nowadays--the independence form the offline trajectories\cite{yang2023cajun,li2023robust,ding2023safe}. However, their need of massive training set and iteration are somehow alike to the pre-optimization process of the trajectories, and the posture control is achieved by imitation learning or parameter tuning. In this paper, we extract significant parameters from the full-scale model to design various models for jump motion optimization. Our goal is to balance the solver's efficiency and the performance of the trajectories. Simultaneously, the optimization of inertia shaping trajectory cooperating with launching trajectory optimization is discussed.

Notably, jumping entails a prolonged period of ballistic flight where angular and horizontal linear momentum remain conserved. With such conservation, the intimate connection between the CAM and composite rigid body (CRB) inertia \cite{walker1982efficient} shaping makes inertia shaping the only way to control the posture during the flight phase. Inertia shaping for humanoid robots has been developed for applications like falling protection\cite{nagarajan2010generalized,yun2009safe} and kick motion\cite{ficht2018online,ficht2020fast}. Similar idea can be utilized on the balance control of humanoid robots\cite{nenchev2018momentum,siravuru2018reaction,garcia2021mpc} and animation generation\cite{zordan2014control,kwon2017momentum}. Zhou etc. \cite{zhou2022momentum} introduced an optimization framework including inertia characters for quadrupedal jumping locomotion. But for humanoid robots, the affect of the trunk's posture to the landing stability can be more significant. As previously mentioned, designing a complex whole-body optimization problem using full-scaled dynamic model can be a potential solution. However, this approach requires substantial computational resources and imposes a heavy burden on tuning. In this paper, we focus on optimizing the inconsistency of the posture and the CAM with high efficiency.

In this paper, we introduce an offline trajectory optimization framework (Fig. \ref{Fig Overview of the framework}), focusing on the matching of robot's aerial motion and inertia shaping. The optimization framework includes three parts and three corresponding dynamic models to improve the optimization speed (the whole optimization process takes less than 10 seconds) while maintaining the accuracy and practicality. The first part is a SRMP model optimization, which generates desired momentum inertia trajectories firstly. Then, the second part maps those trajectories to joint space, generating coarse whole body trajectory. Lastly, the whole body optimization part is fed with trajectories generated by previous parts to output refined trajectory. The main contribution of this paper are as follows:
\begin{enumerate}
	\item The leverage of the posture and the CAM is explored for the launching phase optimization of jump motion by introducing the SRMP model in the launching phase optimization.
	\item The inertia, posture and CAM are collaboratively optimized in this framework to generate an inertia shaping motion during the flight phase, enabling the robot to achieve an ideal posture and foot placement upon touchdown.
	\item The 3-step optimization framework is capable of generating the desired trajectory around 10 seconds, making it feasible to deploy the framework in an online setting.
\end{enumerate}

The remainder of this paper is organized as follows. Section \ref{Sec Modeling of  the robot dynamics for the three-step optimization} introduces the modeling process of the three dynamic models corresponding to the three-step optimization. Section \ref{Sec SRMP Model Optimization} introduces part I of the framework including the inertia and momentum optimization based on the SRMP model. Section \ref{Sec Joint Space Mapping Optimization} constructs part II which maps the SRMP trajectory to coarse joint trajectories. Section \ref{Sec Whole Body Optimization} introduces the final part of the framework which is fed with up-level trajectories and generates the whole body trajectory. Section \ref{Sec Applications on Robot} illustrates the simulation results. Section \ref{Sec Conclusion} summarizes this paper and induces the future work.

\section{Modeling of the robot dynamics for the three-step optimization}
\label{Sec Modeling of the robot dynamics for the three-step optimization}
In this paper, three distinct dynamic models are employed, with each part of the optimization framework utilizing one model. To facilitate comprehension of the theory and the relationship between these models and the optimization framework, we first introduce the dynamic modeling of the robot platform. Subsequently, models are introduced in accordance with the sequence of the optimization framework's three parts.

\begin{figure}[]
	\centering
	\subfigure[]
	{
		\centering
		\includegraphics[width=26mm,height=41mm]{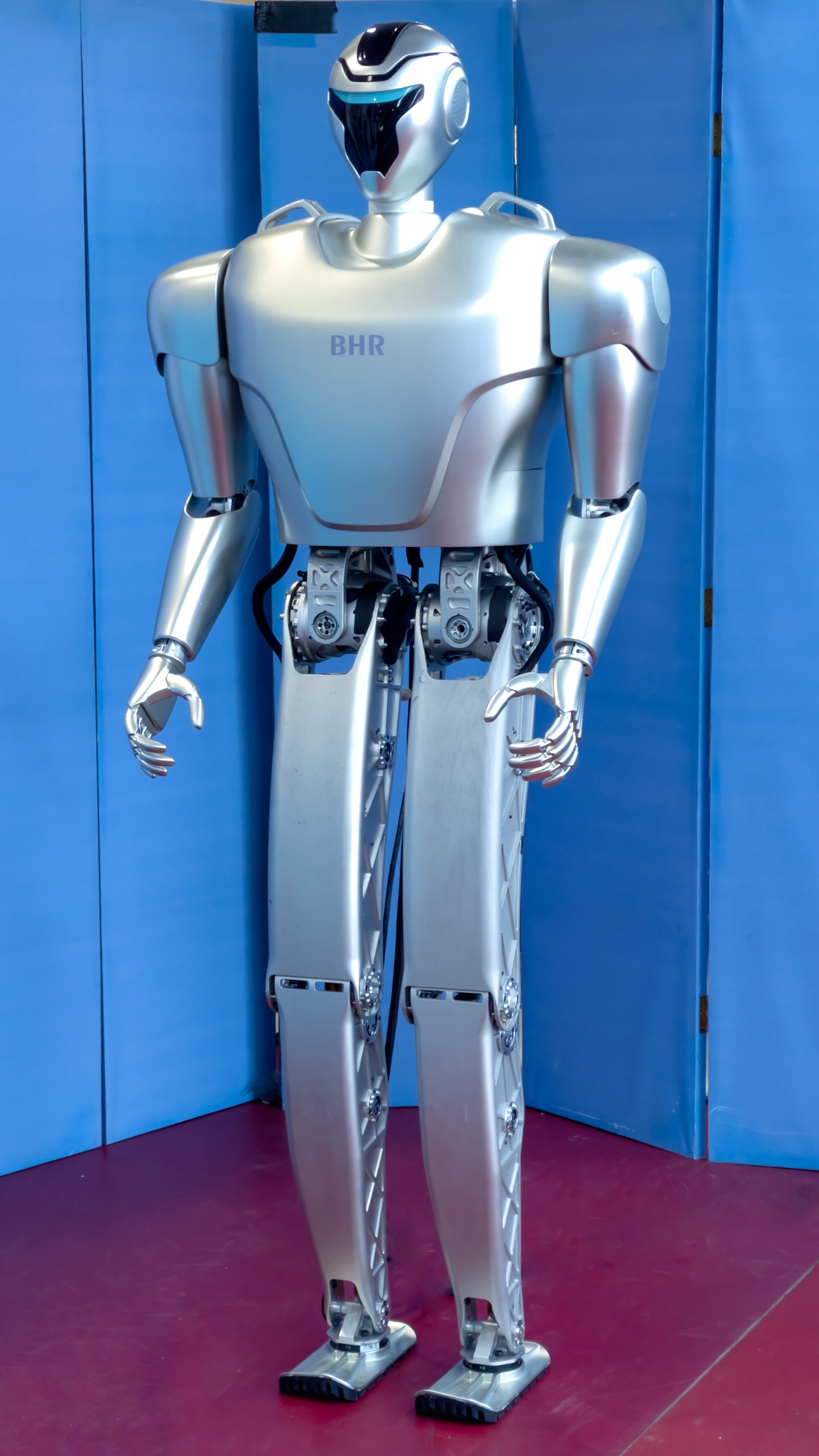}
	}
	\subfigure[]
	{
		\centering
		\includegraphics[scale=0.20]{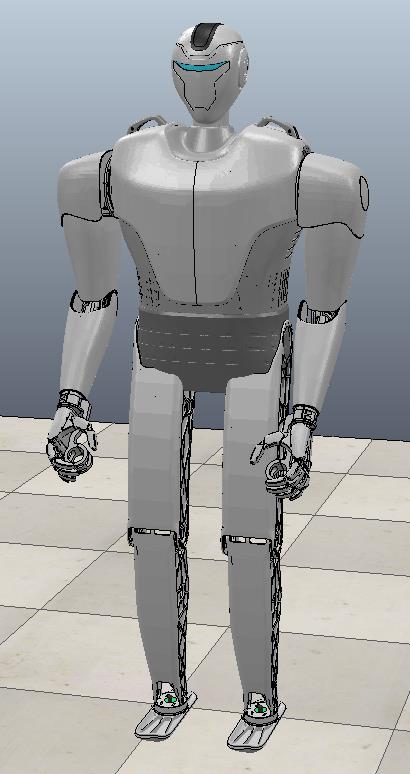}
	}
	\subfigure[]
	{
		\centering
		\includegraphics[scale=0.31]{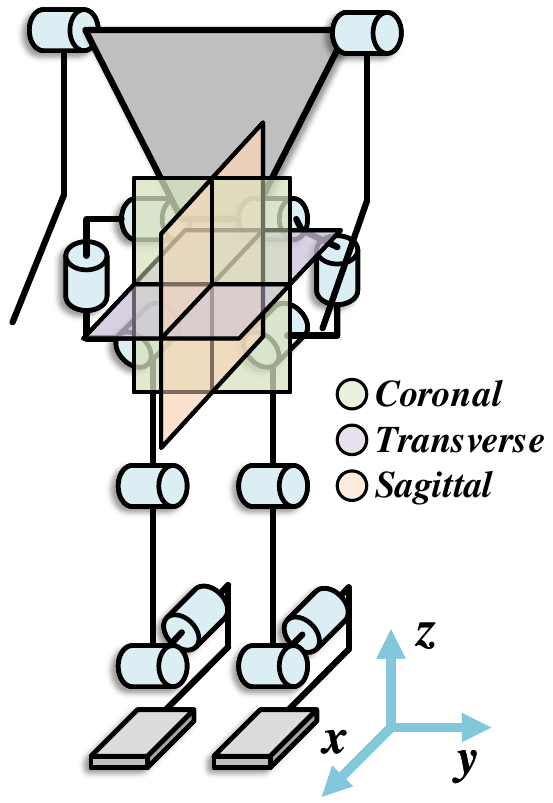}
	}
	\caption{The robot platform used in this paper: (a) actual platform, (b) simulation model, (c) simplified link model}
	\label{Fig Application Robot Model}
\end{figure}

\subsection{SRMP Model}
\label{Sec SRMP Model}
The core of the RMP model is separating the CoM to barbells and adding extra rotational degree of freedom (DOF). Goswami \cite{lee2007reaction,sanyal2014dynamics} etc. introduced a 3 dimensional (3D) RMP model, which includes three barbells and each rotates in a dimension, respectively. Yet in this paper, the jump motion and inertia shaping are performed mainly in the sagittal dimension. Therefore, the 3D RMP model is simplified to 2 dimensional (2D) in the offline optimization framework to generate inertia shaping around the $y$ axis, and the supporting point of the pendulum is fixed to the ground to obtain static dynamic of the model. Therefore, the model utilized in this paper is named SRMP model. In Fig. \ref{Fig Illustration of the RMP}, the SRMP model is illustrated briefly. Additionally, In \cite{lee2007reaction}, the author drew an ellipse to demonstrate the inertia shaping. Accordingly, the demonstrative ellipse is also drawn in light green in this paper.

The configuration parameters of the SRMP model are defined as
\begin{equation}
	\bm{\vartheta}=[\theta_{l};\phi;\theta_{pl};\varphi]
\end{equation}
and the static dynamic model is derived by the Lagrange equation
\begin{equation}
	\label{Eq SRMP dynamic equation}
	\bm{M}^{sr}(\bm{\vartheta})\ddot{\bm{\vartheta}}+\bm{H}^{sr}(\bm{\vartheta},\dot{\bm{\vartheta}})=\bm{F}^{sr}
\end{equation}
where the superscript $^{sr}$ is used to represent the \textquotedblleft SRMP\textquotedblright; $\bm{M}^{sr}$ denotes the mass and inertia matrix; $\bm{H}^{sr}$ is the Coriolis and gravity matrix; $\bm{F}^{sr}$ stands for the joint force/torque:
\begin{equation}
	\bm{F}^{sr} = [\tau_l;F_{\phi};\tau_{pl};F_{\varphi}]
\end{equation}

Based on the configuration of the SRMP model, the total inertia of the model $\rho_{sr}$ can derived as:
\begin{equation}
	\rho_{sr}=2m_p\varphi^2
\end{equation}
where $m_p$ denotes the mass of the barbell's endpoint.

\subsection{Full-DOF Dynamic Model}
\label{Sec Full-DOF Dynamic Model}
The whole-body robot model (Fig. \ref{Fig Application Robot Model}(c)) is utilized in Sec. \ref{Sec Joint Space Mapping Optimization} by QP solvers to obtain guesses of whole-body trajectories rapidly. The robot is modeled as a 9-link dynamic system in 3D, which has 20 DOF including a 6-DOF floating base, 1-DOF shoulders, 3-DOF hips, 1-DOF knees, and 2-DOF ankles. The physical parameters are the same with the platform utilized in \cite{qi2023vertical}. The configuration parameters are defined as
\begin{equation}
	\bm{\theta}^{js}=[x; y; z; q_{roll}; q_{pitch}; q_{yaw}; \bm{q}^{js}]
\end{equation}
where $x$, $y$, and $z$ are the floating base positions fixed to the top of the trunk. $q_{roll}$, $q_{pitch}$, and $q_{yaw}$ are the body posture angles in the roll, pitch and yaw axes in the world frame. $\bm{q}^{js}$ is the joint angle vector. The robot dynamics can be derived by the Lagrange equation
\begin{equation}
	\label{Eq Joint Space Mapping Dynamic Equation}
	\bm{M}^{js}(\bm{\theta}^{js})\ddot{\bm{\theta}}^{js}+\bm{H}^{js}(\bm{\theta}^{js},\dot{\bm{\theta}}^{js})=\bm{S}^T\bm{\tau}^{js}+(\bm{J}^{js})^T\bm{F}_{ext}^{js}
\end{equation}
where the superscript $^{js}$ is used to represent the \textquotedblleft Joint Space Mapping\textquotedblright; $\bm{M}^{js}$ and $\bm{H}^{js}$ stand for the same meaning with $\bm{M}^{sr}$ and $\bm{H}^{sr}$ in Sec. \ref{Sec SRMP Model}, respectively. $\bm{S}^T$ denotes the selector matrix, which depends on the available actuator; $\bm{\tau}$ denotes the joint torque vector; $\bm{F}^{js}_{ext}\in R^{12\times 1}$ denotes the external force/torque matrix, and each foot has six degrees of external force/torque (three forces and three torques), respectively; and $\bm{J}^{js}$ denotes the Jacobian matrix corresponding to $\bm{F}^{js}_{ext}$.

\subsection{Symmetrical-Body Dynamic Model}
\label{Sec Symmetrical-Body Dynamic Model}
The motion of the two legs is intended to be symmetric during the jump motion to ensure a stable landing. Therefore, prioritizing solver efficiency, the robot is reconfigured as a single-legged 5-link system to circumvent the symmetric problem and reduce solving time. The physical parameters of the single leg are an aggregation of those of the two legs, encompassing mass and inertia. Consequently, the robot with two legs can execute the jump motion as anticipated by utilizing the solver output symmetrically on each leg. The configuration parameter $\bm{\theta}^{wb}$ is defined as
\begin{equation}
	\bm{\theta}^{wb} = [x;z;\theta_{pitch};\bm{q}^{wb}]
\end{equation}
where the floating base only contains linear DOFs along the $x$ and $z$ coordinates and angular DOF around the pitch axe. The joint angle vector $\bm{q}^{wb}$ contains three leg DOFs and one arm DOF, all of which are rotational DOF around the pitch axe. Then, the dynamic equation can be derived in accordance to previous models:
\begin{equation}
	\bm{M}^{wb}(\bm{\theta}^{wb})\ddot{\bm{\theta}}^{wb}+\bm{H}^{wb}(\bm{\theta}^{wb},\dot{\bm{\theta}}^{wb})=\bm{S}^T\bm{\tau}^{wb}+(\bm{J}^{wb})^T\bm{F}_{ext}^{wb}
\end{equation}
where the superscript $^{wb}$ is used to represent the \textquotedblleft Whole Body\textquotedblright; the external force/torque vector $\bm{F}_{ext}^{wb}$ contains only force along the $x$ and $z$ coordinates and torque around the pitch axe.

\begin{figure}
	\centering
	\includegraphics[scale=0.5]{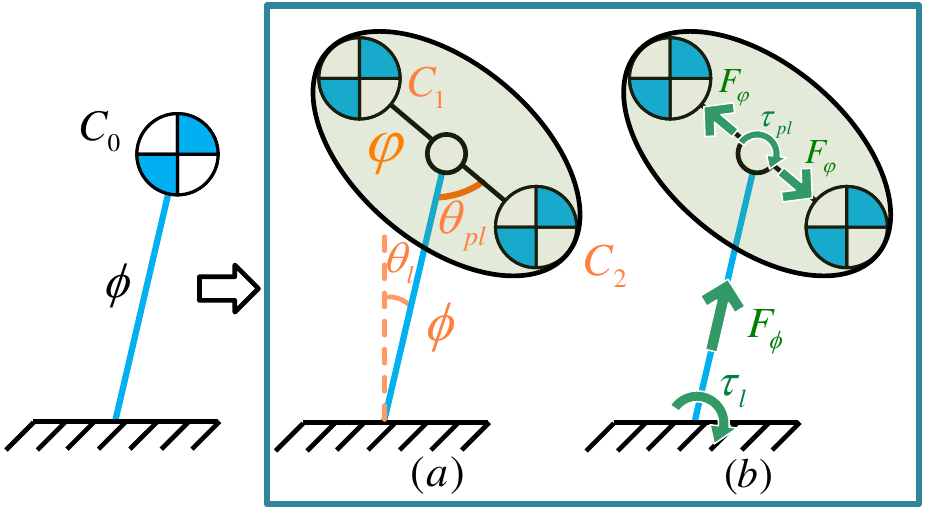}
	\caption{Diagram of the reaction mass pendulum (RMP) model. The CoM of the inverted pendulum (left-hand side) $C_0$ is divided into a barbell on the sagittal dimension, and the mass is separated to two pieces evenly ($C_1$ and $C_2$) on the endpoint of the barbell. The geometrical characters and dynamical characters are illustrated in (a) and (b), respectively. The length of the pendulum's support segment is represented by $\phi$, and the radius of the barbell is represented by $\varphi$.}
	\label{Fig Illustration of the RMP}
\end{figure}

\section{SRMP Model Optimization}
\label{Sec SRMP Model Optimization}
As mentioned in Sec. \ref{Sec SRMP Model}, the robot is simplified as a SRMP model. As shown in Fig. \ref{Illustration of jump with RMP}, we consider a jump motion as a symmetrical process that contains identical launching and landing angle. Such a symmetry benefits the robot in absorbing the momentum while landing, because the robot's CoM has enough space to decelerate. Additionally, the SRMP model contains a DOF of the ellipse's rotation, which is similar to the torso's posture angle by considering the support rod as the robot's leg and the ellipse as the torso. Therefore, it is possible to optimize the launching posture by utilizing the SRMP model.

\begin{figure}
	\centering
	\includegraphics[scale=0.11]{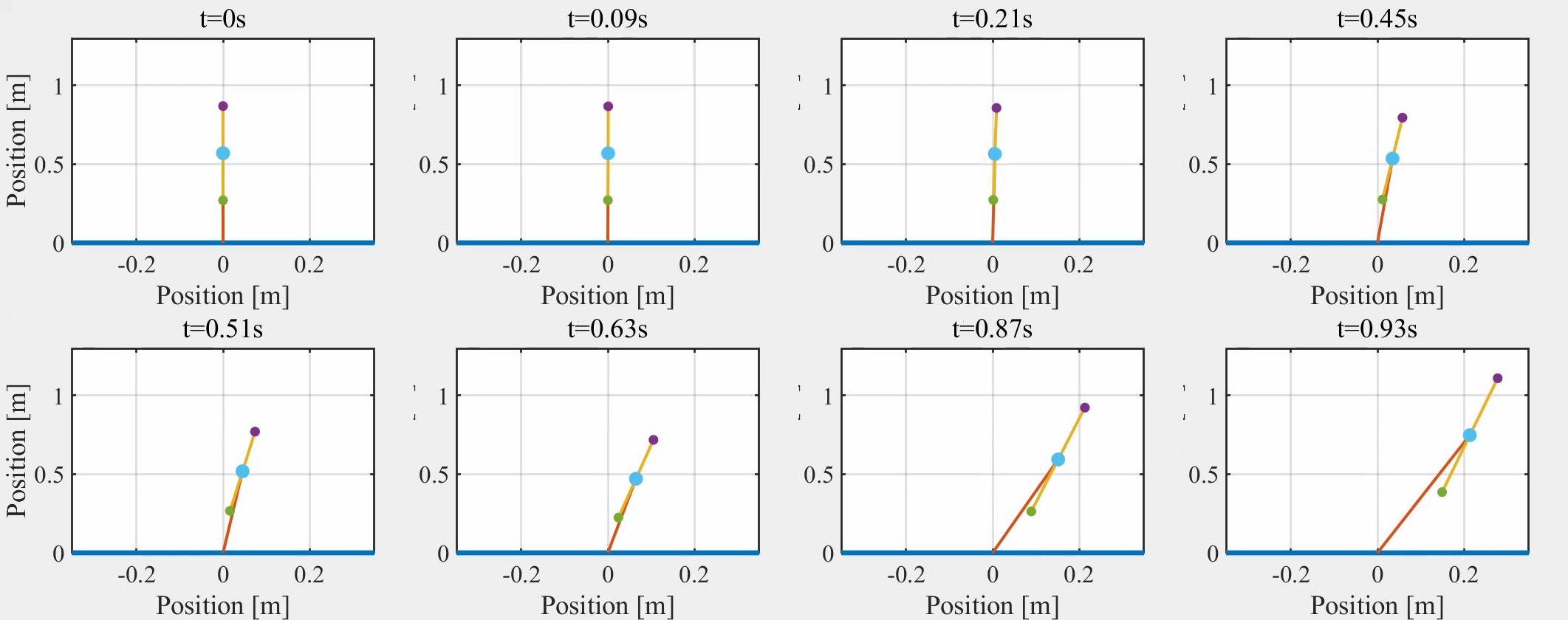}
	\caption{Illustration of the optimized SRMP trajectory during the launching phase from Sec. \ref{Sec SRMP Model Optimization}.}
	\label{Fig Illustration of the SRMP trajectory}
\end{figure}

By setting the height and the distance of a jump as $h$ and $l$, respectively, we can firstly separate $l$ into distance on the land $l_g$ and distance in the air $l_f$, which are uncertain before the optimization. Then, we set the gravitational parameter as $g=9.8$ and derive expected CoM's velocity in the $z$-axis $v_z^{exp}=\sqrt{2gh}$. The duration time of the flight phase can also be solved by $t^{sf}=2v_z^{exp}/g$. We utilize the optimization strategy in \cite{qi2023vertical} to generate landing trajectory. A simple illustration of the optimized trajectory during the launching phase is depicted by Fig. \ref{Fig Illustration of the SRMP trajectory}.

\subsection{Launching Phase Optimization}
\label{Sec SRMP Model Optimization Launching}
\subsubsection{Problem Formulation}
In this paper, we focus on 2D jump motions like forward jump and backflip, and to accelerate the optimization, we only optimize momentum parameters on the sagittal plane. 3D jump motion optimizations can be analogized based on it. According to the RMP model in Fig. \ref{Fig Illustration of the RMP}, the state parameters $\bm{s}^{sl}$ (where the superscript $^{sl}$ stands for \textquotedblleft SRMP launching\textquotedblright) can be listed:
\begin{equation}
	\bm{s}^{sl}=(\theta_l, \phi, \theta_{pl}, \varphi, \omega_l, \dot{\phi}, \omega_{pl}, \dot{\varphi}, \dot{\omega}_l, \ddot{\phi}, \dot{\omega}_{pl}, \ddot{\varphi})
\end{equation}
To smooth the trajectory, jerk of the geometrical characters of the RMP is set as the control parameters, and the dynamical characters are set as constraints. Therefore, the control parameters $\bm{u}$ can be listed:
\begin{equation}
	\bm{u}^{sl}=(\ddot{\omega}_l, \dddot{\phi}, \ddot{\omega}_{pl}, \dddot{\varphi})
\end{equation}

By setting the total mass of the robot as $m=2m_p$, the boundary constraints are deployed for $\bm{s}^{sl}$, $\bm{u}^{sl}$ and $\bm{F}^{sr}$ first to limit their minimum and maximum thresholds. These bounds are manually designed and do not need to be highly precise, as subsequent constraints, such as end constraints and inertia constraints, will further limit these parameters.


\begin{figure}
	\centering
	\includegraphics[scale=0.55]{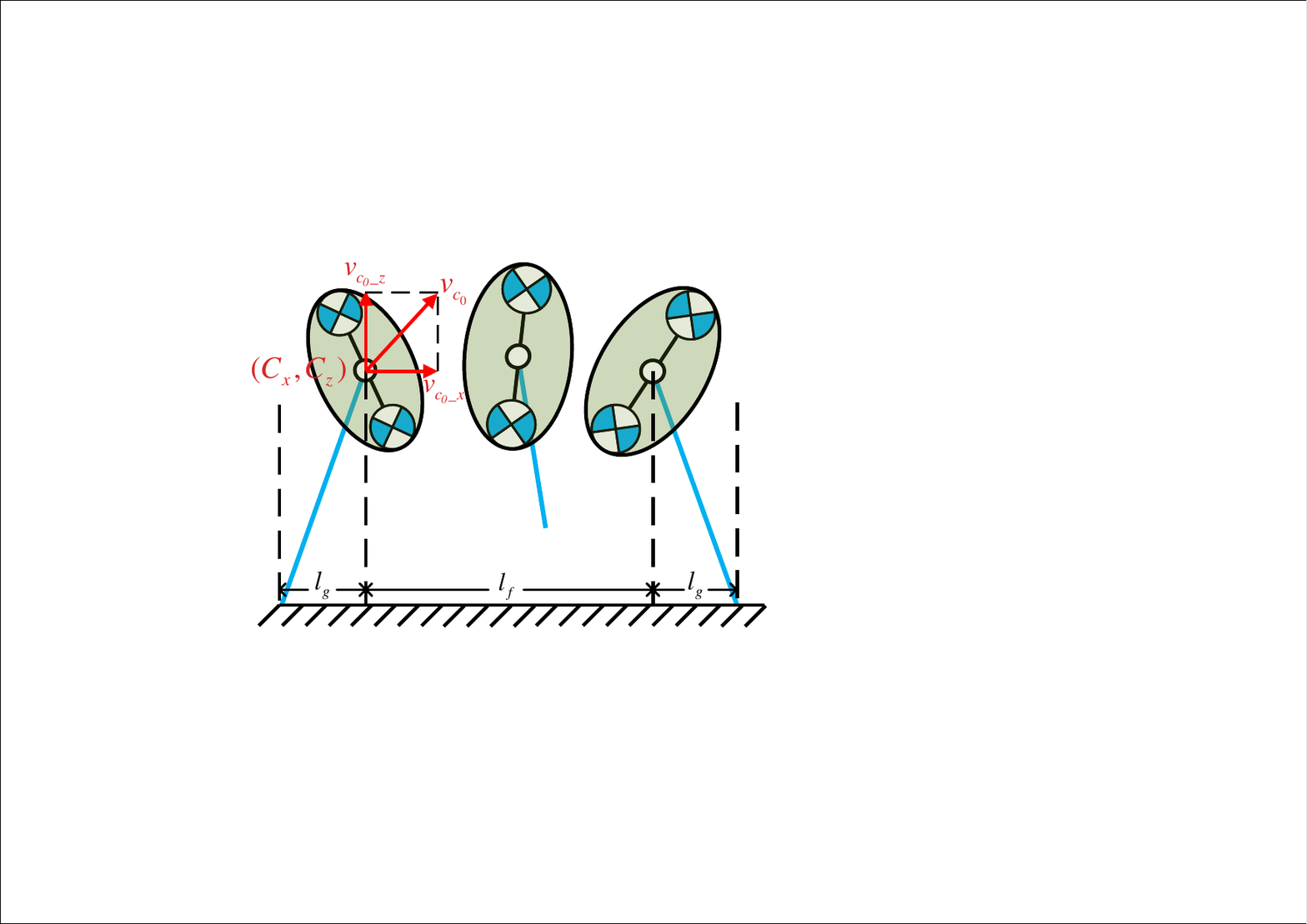}
	\caption{Diagram of the jump motion with the inverted pendulum model. $C_x$ and $C_z$ represent the position of the CoM in the x-axis and z-axis, respectively. $v_{c_0\_x}$ and $v_{c_0\_z}$ represent the velocity of the CoM in the x-axis and z-axis, respectively, at the same moment. $\theta$ represent the launching and landing angle. $l_g$ and $l_f$ are distance in the x-axis distinguished by whether the robot's feet touch the ground.}
	\label{Illustration of jump with RMP}
\end{figure}

\subsubsection{Cost Function}
\label{Launching Cost Function}
The items inside the cost function can be categorized into three groups: \emph{control smoothness}, \emph{state smoothness} and \emph{final state target}.

To generate a smooth motion, the control inputs (the angular jerk and derivations of linear jerk) are minimized:
\begin{equation}
	\label{Eq Opt launching cost function}
	\sum_{k=0}^{N_t^{sl}}\|\bm{u}^{sl}[k]\|^2_{W_u^{sl}}
\end{equation}
where $N_t^{sl}$ denotes the number of the timestep in the launching phase, and $k$ denotes the serial number of the timestep. $W_u^{sl}$ denotes the weight matrix.

Despite the jump motion is intense and highly dynamic, the motion during the launching phase is expected to be mitigatory and smooth. So that the robot won't perform exaggerated motion and the burden of the hardware system can be relieved. Therefore, the posture at the end of the launching phase and the velocity of the CoM $v_{c_0}$ are minimized:
\begin{equation}
	\|\theta_l^{end}+\theta_{pl}^{end}\|^2+\sum_{k=0}^{N_t^{sl}}\|v_{c_0}[k]\|_{W_v^{sl}}^2
\end{equation}

Apart from the linear momentum, the trajectory of the angular momentum is also supposed to be smooth and optimized. For the RMP model, its mass is concentrated on the endpoints of the barbell. Thus, the barbell's momentum equals to the system's momentum, and the CAM $L$ can be formed as:
\begin{equation}
	L=2m_p\varphi^2(\omega_l+\omega_{pl})
\end{equation}
and the cost function can be formed as:
\begin{equation}
	\|L^{end}\|^2_{W_L^{sl}}+\sum_{k=0}^{N_t^{sl}}\|\dot{L}[k]\|_{W_L^{sl}}^2
\end{equation}

Posture also affects the stability of the robot significantly, especially for a forward jump. Unexpected posture at the moment of taking off, such as leaning back too much, can lead to unstable landing. For the RMP model in this paper, $\theta_{pl}$ represents the posture of the robot. Therefore, $\theta_{pl}$ is included in the cost function:
\begin{equation}
	\sum_{k=0}^{N_t^{sl}}\|\theta_{pl}[k]\|^2_{W_{pl}^{sl}}
\end{equation}
Meanwhile, the weight for $\dddot{\theta}_{pl}$ in Eq. \ref{Eq Opt launching cost function} is set larger than other parameters.

Since the motion of the launching phase is also supposed to be effective and swift, the total time of the launching phase $t^{sl}$ is taken into the consideration:
\begin{equation}
	\|t^{sl}\|^2_{W_t^{sl}}
\end{equation}

\subsubsection{End Constraints}
From Fig. \ref{Illustration of jump with RMP} we can see that the jump distance is a combination of lean motion and flying in the air. $l_g$ actually equals to the CoM's position at the end of the launching phase $p_{c_0\_x}^{end}$. Following this concept, the jump distance constraint is set:
\begin{equation}
	2p_{c_0\_x}^{end}+v_{c_0\_x}^{end}t_f=l
\end{equation}

To ensure that the robot enters the flight phase smoothly and reaches desired velocity at that moment, corresponding constraints are made:
\begin{equation}
	\left\{
	\begin{array}{l}
		v_{c_0\_z}^{end}=v_z^{exp}   \\
		F_{l}^{end}=\tau_{l}^{end}=0 \\
	\end{array}
	\right.
\end{equation}

In Sec. \ref{Launching Cost Function}, the inertia parameter of the system is derived. Thereout, the inertia is constrained to a proper range respecting to the actual robot. The derivation of the CAM at the end of the launching phase is also constrained to smooth the trajectory:
\begin{equation}
	\left\{
	\begin{array}{l}
		\rho_{sr}^{min}\leq 2m_p(\varphi^{end})^2\leq \rho_{sr}^{max} \\
		2m_p(\varphi^{end})^2(\dot{\omega}_l+\dot{\omega}_{pl})+4m_p\varphi^{end}\dot{\varphi}^{end}(\omega_l+\omega_{pl})=0
	\end{array}
	\right.
\end{equation}

\subsubsection{Posture Constraint}
As the Fig. \ref{Fig Illustration of the RMP} illustrated, the posture of the dumbbell can be seen as the posture of the upper trunk of the robot. Therefore, we constrain its posture at the endpoint to avoid unexpected posture at the moment of takeoff and obtain proper CAM simultaneously.
\begin{equation}
	\theta_{pos}^{min}\leq \theta_{l}+\theta_{pl}\leq \theta_{pos}^{max}
\end{equation}

\subsubsection{CoP Constraints}
On the one hand, generating proper CAM is significant for performing a forward jump, but it's not promising to generate the CAM trajectory in isolation because the linear momentum and the angular momentum are related by the CoP. From the equation below we can see it more clearly\cite{tajima2006motion}:
\begin{equation}
	\dot{L}=-mX_{CoP}(\ddot{p}_{c_0\_z}+g)+mp_{c_0\_x}(\ddot{p}_{c_0\_z}+g)-mp_{c_0\_z}\ddot{p}_{c_0\_x}
	\label{CoP equation}
\end{equation}
where $X_{CoP}$ is the position of the CoP in the x-axis. The equation reveals the relationship between the CoP and the derivations of the linear momentum and the angular momentum. In other words, the angular momentum is restricted by the linear momentum and the support region.

On the other hand, the CoP is supposed to be restricted in the support region so that the robot can perform the motion stably in without exaggerated foot tilting. But the RMP model utilized in this phase is considered as a fixed-base model, which provides unlimited supported force/torque $F_l,\tau_l$ to the RMP. Therefore, we set the support region as a constraint to ensure the rationality and executability of the optimized CAM trajectory by transforming Eq. \ref{CoP equation}:
\begin{equation}
	X_{CoP}^{min}\leq p_{c_0\_x}-\frac{mp_{c_0\_z}\ddot{p}_{c_0\_x}+\dot{L}}{m(\ddot{p}_{c_0\_z}+g)}\leq X_{CoP}^{max}
\end{equation}
Additionally, the initial position of the CoP is also restricted to equal to the initial $x$-axis position of the CoM.

\subsubsection{Inertia and Angular Momentum Constraints}
Apart from the end constraints of the inertia and CAM, path constraints are also necessary for them to limit them in a proper range during the launching phase and avoid unexpected motion:
\begin{equation}
	\left\{
	\begin{array}{l}
		\rho_{sr}^{min}\leq 2m_p\varphi^2\leq \rho_{sr}^{max}                       \\
		L^{min}\leq 2m_p\varphi^2(\omega_l+\omega_{pl})\leq L^{max} \\
	\end{array}
	\right.
\end{equation}

\subsection{Flight Phase Optimization}
\subsubsection{Problem Formulation}
When the robot takes off, the gravitational force is the only external force acting on the robot, making the angular momentum and horizontal linear momentum constant. In spite of this, the rotational velocity around the CoM, which changes with the inertia shaping, is still controllable. By planning and controlling the rotational velocity, we can adjust the landing posture to ensure the stability of landing, making the control framework more robust. Therefore, inertia shaping is the core parameter to be optimized in the flight phase.

Moreover, due to the massless characteristics of the RMP's support rod, it is no longer appropriate to consider the rod as the robot's lower body. After all, the massless rob is unable to affect the rotation of the ellipse in the flight phase. Therefore, we focus on the inertia shaping of the ellipse in this phase, and the state parameters $\bm{s}^{sf}$ and control parameters $\bm{u}^{sf}$ (where the superscript $^{sf}$ stands for \textquotedblleft SRMP flight\textquotedblright) become:
\begin{equation}
	\bm{s}^{sf}=(\theta_l, \rho_{total}^{R}, \dot{\rho}_{total}^{R}, \ddot{\rho}_{total}^{R}),\quad \bm{u}^{sf}=\dddot{\rho}_{total}^{R}
\end{equation}
where $\theta_l$, which denotes the launching angle in the launching phase, denotes the angle between the line connected by CoM and foot and the vertical line in this section. The inertia parameter of the whole-body dynamic model $\rho_{total}$ is divided into two parts: relative inertia $\rho^{R}_{total}$ (the so-called spatial inertia\cite{orin2013centroidal}) and internal inertia $\rho^{I}_{total}$. Then, the corresponding boundary constraints are deployed to constrain $\bm{s}^{sf}$ and $\bm{u}^{sf}$.
\subsubsection{Cost Function}
The items inside the cost function are designed to smooth the output. $\ddot{\rho}_{total}^{R}$ and $\bm{u}^{sf}$ are minimized:
\begin{equation}
	\sum_{k=0}^{N_t^{sf}}\|\ddot{\rho}^R_{total}[k]\|^2_{W^{sf}_I}+ \sum_{k=0}^{N_t^{sf}}\|(\bm{u}^{sf}[k])\|^2_{W^{sf}_u}
\end{equation}
\subsubsection{State Constraints}
In the optimization, we consider $\theta_l$ as a varying parameter to constraint the inertia state parameter. In this case, the angular velocity $\omega_l$ changes with the inertia shaping:
\begin{equation}
	\omega_l=L/(\rho^I_{total}+\rho^R_{total})
\end{equation}
and eventually affects the landing posture. For jump motions like vertical jump or forward jump, the landing angle $\theta_l^{end}$ can be set as symmetric to the launching angle $\theta_l^{init}$, To make sure the robot is able to absorb the horizontal momentum and achieve desired rotation $\theta^{tar}$ during the flight phase (a backflip for example), we set:
\begin{equation}
	\theta^{end}_l + \theta^{init}_l=\theta^{tar}
\end{equation}
which makes the landing posture symmetric with the launching posture, giving the robot enough space to decelerate the CoM. Additionally, the duration of the flight phase can be derived by $t_{sf}=2v_z^{exp}/g$.

\subsubsection{Link Constraints}
Since the optimization is separated into two phases with different state and control parameters, it is important to ensure the continuity of some parameters, such as the robot's inertia. Therefore, the link constraints are established:
\begin{equation}
	\left\{
	\begin{array}{l}
		\theta_l^{sl\_end}=\theta_l^{init}    \\
		\rho_{sl}^{end}=\rho^{R\_init}_{total}+\rho^{I\_init}_{total}   \\
		\dot{\rho}_{sl}^{end}=\dot{\rho}^{R\_init}_{total}   \\
		\ddot{\rho}_{sl}^{end}=\ddot{\rho}^{R\_init}_{total} \\
	\end{array}
	\right.
\end{equation}
where $\theta_l^{sl\_end}$ denotes the endpoint value of $\theta_l$ in Sec. \ref{Sec SRMP Model Optimization Launching}'s result.

\section{Joint Space Mapping Optimization}
\label{Sec Joint Space Mapping Optimization}
After the SRMP optimization, trajectories of momentum and inertia are acquired. However, there is one crucial step remaining before the final whole-body optimization - the preparation of the optimization guess. It is widely recognized that the quality of the guess significantly impacts the speed and effectiveness of optimization algorithms. In this section, we map the momentum and inertia trajectories to the robot's joint space using QP solvers, utilizing the results as the initial guess for the optimization in Sec. \ref{Sec Whole Body Optimization}. While running Sec. \ref{Sec Whole Body Optimization} with more tolerances could also generate an initial guess, the guess obtained in this section is more efficient. Specifically, the joints' motion is constrained to closely track the momentum and inertia trajectories with minimal tolerances, resulting in a trajectory close to an ideal jump. With such an initial guess, the optimization process in Sec. \ref{Sec Whole Body Optimization} can be further accelerated.

\begin{figure}
	\centering
	\includegraphics[width=70mm,height=60mm]{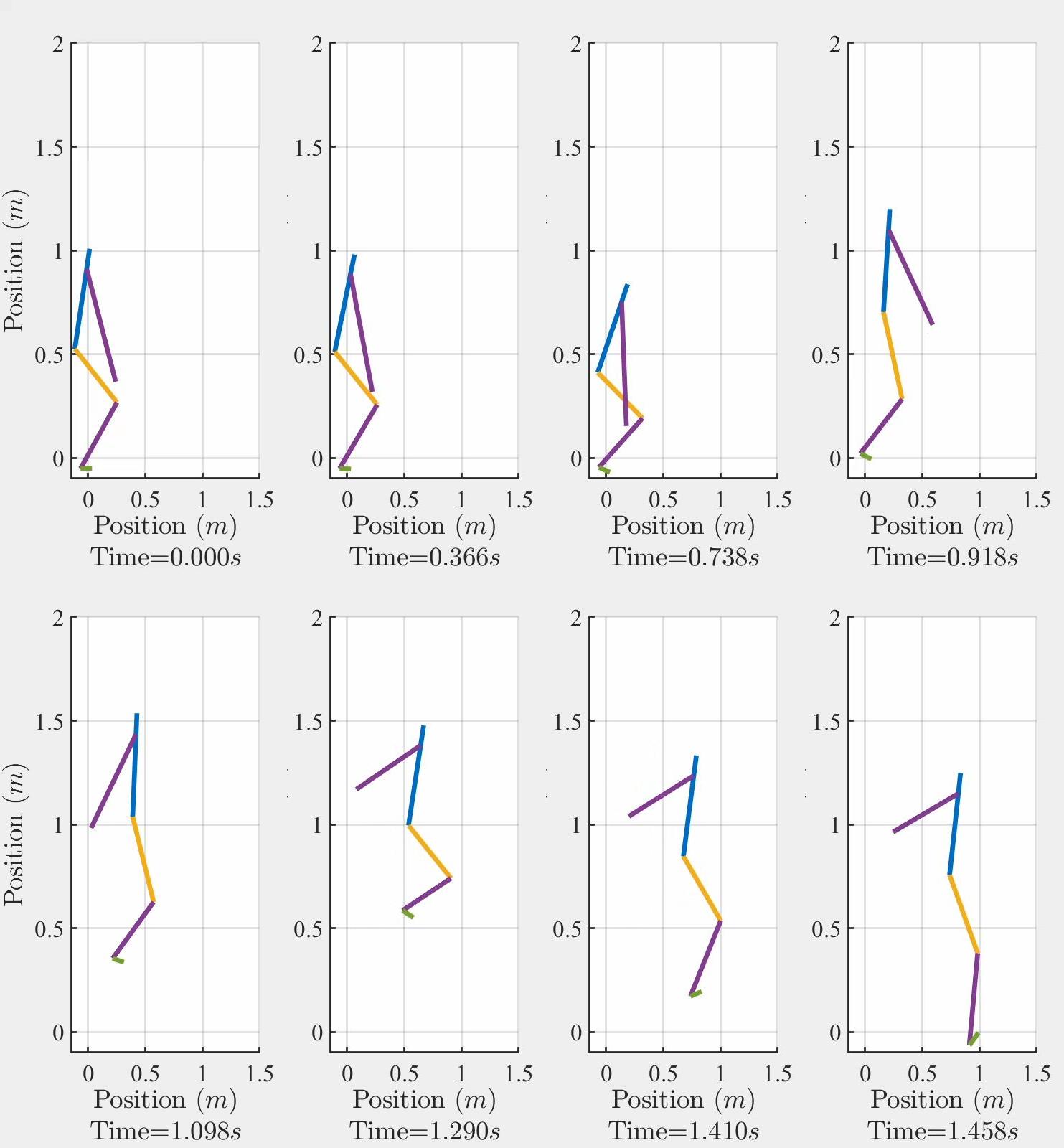}	
	\caption{Illustration of the optimized body trajectory from Sec. \ref{Sec Joint Space Mapping Optimization}. This brief optimization lacks constraints on posture; hence, an unexpected posture is observed when the robot touches the ground. However, this has no adverse effect on the ultimate results, as it is solely used as the initial guess for the whole-body optimization.}
	\label{Fig JSP Snapshot}
\end{figure}

As introduced in Sec. \ref{Sec Full-DOF Dynamic Model}, the robot is modeled as a 9-link system. Similar to Sec. \ref{Sec SRMP Model Optimization}, this optimization is divided into parts - launching phase and flight phase, and the QP is constructed differently in each phase. Results of this section are illustrated by a snapshot in Fig. \ref{Fig JSP Snapshot}.

\subsection{Launching Phase}
During this phase, the ground reaction force/torque accelerates the robot until it attains the desired momentum, and Eq. \ref{Eq Joint Space Mapping Dynamic Equation} offers a thorough and elucidating insight into the fundamental principles. Consequently, the floating base dynamics component (top three rows) of Eq. \ref{Eq Joint Space Mapping Dynamic Equation} is expected to be incorporated into the equality constraints of the QP problem. The dynamics constraint is illustrated as follows:
\begin{equation}
	\label{Eq Floating Base Dynamics Part of JSM Launching}
	\tilde{\bm{M}}^{js}\ddot{\bm{\theta}}^{js}+\tilde{\bm{H}}^{js}=(\tilde{\bm{J}}^{js})^T\bm{F}_{ext}^{js}
\end{equation}
where $\tilde{\bm{M}}^{js}$ and $\tilde{\bm{H}}^{js}$ denote the top three rows of $\bm{M}^{js}$ and $\bm{H}^{js}$, respectively; $\tilde{\bm{J}}^{js}$ denotes the left three columns of $\bm{J}^{js}$; $\bm{F}_{ext}^{js}$ denotes the external force and torque.

In accordance with Sec. \ref{Sec SRMP Model Optimization}, the momentum trajectories are obtained. In \cite{qi2023vertical}, the relationship the relationship between the ground reaction force/torque, gravity, and the acceleration of the CoM has been derived (Eq. 29, 32 and 33 of \cite{qi2023vertical}). In this paper, we add the derivation of the CAM $\dot{L}$ to them and derive a simplified version of Eq. 29 of \cite{qi2023vertical}:
\begin{equation}
	\mathcal{A}_f\bm{F}_{ext}^{js}=\mathcal{B}_f
\end{equation}
where $\mathcal{A}_f$ denotes the kinematic matrix of external force and torque, revealing the relationship between $\bm{F}_{ext}$ and $\mathcal{B}_f$, which denotes the derivation of linear momentum and angular momentum.

As for the joint acceleration, the centroidal dynamics equation and tiptoe constraint introduced in \cite{qi2023vertical} are utilized:
\begin{equation}
	\label{Eq Joint Acceleration Equation in Launching}
	\begin{bmatrix}
		A_c \\ \bm{J}^{js}
	\end{bmatrix}
	\ddot{\bm{\theta}}^{js}=
	\begin{bmatrix}
		\ddot{c}^{ref}-B_c \\ -\dot{\bm{J}}^{js}\dot{\bm{\theta}}^{js}+\ddot{p}_{foot}^{js}
	\end{bmatrix}
\end{equation}
where $A_c$ and $B_c$ are derived through the relationship between velocity of the CoM and the joint space, and $p_{foot}^{js}$ denotes the position vector of both tiptoes. In addition to constraints on ground reaction force/torque and joint acceleration, a symmetric constraint on the hip joint is employed to guarantee symmetry in the results of both legs. This is achieved by setting the sum of the accelerations of the right and left hip joints equal to zero.

It's also essential to note that in this optimization, there is no ground providing contact force. Consequently, the robot's body, particularly its heel, has the potential to rotate freely around its tiptoe, as the tiptoe is constrained by Eq. \ref{Eq Joint Acceleration Equation in Launching}. Therefore, apart from the equality constraints, the motion of the heel should be constrained by inequality constraints to prevent it from \textquotedblleft sinking\textquotedblright \, beneath the ground. The formulation of the inequality constraint can be illustrated as follows:
\begin{equation}
	\bm{J}_{heel}^z\ddot{\bm{\theta}}^{js}+\dot{\bm{J}}_{heel}^z\dot{\bm{\theta}}^{js}\geq0
\end{equation}
where $\bm{J}_{heel}^z$ denotes the $z$-axis linear momentum part of the Jacobian matrix located on the heel.

Accordingly, the QP problem is constructed as follows:
\begin{gather}
	\hat{\mathcal{X}}=\arg\min\mathcal{X}^{T}W^{jl}\mathcal{X}+f^{jl}\mathcal{X} \\
	\begin{array}{c}
		s.t.\quad \tilde{\bm{M}}^{js}\ddot{\bm{\theta}}^{js}+\tilde{\bm{H}}^{js}=(\tilde{\bm{J}}^{js})^T\bm{F}^{js}_{ext} \\
		\ddot{\theta}_{Rhip}+\ddot{\theta}_{Lhip}=0                                                                  \\
		\bm{J}^{js}\ddot{\bm{\theta}^{js}}=-\dot{\bm{J}}^{js}\dot{\bm{\theta}}^{js}+\ddot{p}_{foot}^{js}             \\
		\bm{J}_{heel}^z\ddot{\bm{\theta}}^{js}+\dot{\bm{J}}_{heel}^z\dot{\bm{\theta}}^{js}\geq0
	\end{array}
\end{gather}
where $\mathcal{X}=[\bm{F}^{js}_{ext};\ddot{\bm{\theta}}^{js}]$ denotes the state vector. $W^{jl}$ and $f^{jl}$ are weight matrix derived as follows:
\begin{gather}
	W^{jl}=
	\begin{bmatrix}
		0 & A_c \\ \mathcal{A}_f & 0
	\end{bmatrix}^{T}
	\begin{bmatrix}
		0 & A_c \\ \mathcal{A}_f & 0
	\end{bmatrix} \\
	f^{jl}=-
	\begin{bmatrix}
		0 & \ddot{c}^{ref}-B_c \\ \mathcal{B}_f & 0
	\end{bmatrix}^{T}
	\begin{bmatrix}
		0 & A_c \\ \mathcal{A}_f & 0
	\end{bmatrix}
\end{gather}

\subsection{Flight Phase}
In the flight phase, ground reaction force/torque no longer exist, leading to the conservation of momentum. Consequently, the floating base dynamics component is rewritten as follows:
\begin{equation}
	\tilde{\bm{M}}^{js}\ddot{\bm{\theta}}^{js}+\tilde{\bm{H}}^{js}=0
\end{equation}
and the momentum and foot constraints in Eq. \ref{Eq Joint Acceleration Equation in Launching} are not needed anymore. Instead, the inertia $I_{total}$ and relative distance $p_{rd}$ between the CoM and foot are constrained in the QP of this phase. Relative equations can be derived as follows:
\begin{equation}
	\begin{bmatrix}
		\bm{J}_\rho^{js} \\ \bm{J}_{rd}^{js}
	\end{bmatrix}
	\ddot{\bm{\theta}}^{js}=
	\begin{bmatrix}
		-\dot{\bm{J}}_\rho^{js}\dot{\bm{\theta}}^{js}+\ddot{\rho}^{ref}_{total} \\ -\dot{\bm{J}}^{js}_{rd}\dot{\bm{{\theta}}}^{js}+\ddot{p}_{rd}
	\end{bmatrix}
\end{equation}
where $\bm{J}_\rho^{js}$ and $\bm{J}_{rd}^{js}$ denote Jacobian matrices of the inertia and relative position, respectively. It is worth noting that $p_{rd}$ is calculated by the trajectory generated by the optimization framework of the flight phase in Sec. \ref{Sec SRMP Model Optimization}. $\ddot{\rho}_{total}^{ref}$ and $\ddot{p}_{rd}$ are derived through PD controllers cooperating with feedforward (reference trajectories).

Accordingly, the QP problem in this phase is constructed as follows:
\begin{equation}
	\hat{\ddot{{\bm{\theta}}}}^{js}=\arg\min(\ddot{{\bm{\theta}}}^{js})^TW^{jf}\ddot{{\bm{\theta}}}^{js}+f^{jf}\ddot{{\bm{\theta}}}^{js}
\end{equation}
\begin{equation}
	\begin{array}{c}
		s.t.\quad \tilde{\bm{M}}^{js}\ddot{\bm{\theta}}^{js}+\tilde{\bm{H}}^{js}=0 \\
		\ddot{\theta}_{Rhip}+\ddot{\theta}_{Lhip}=0                                \\
		J_I^{js}\ddot{\bm{\theta}}^{js}=-\dot{\bm{J}}_I^{js}\dot{\bm{\theta}}^{js}+\ddot{I}_{total}^{ref}
	\end{array}
\end{equation}
$W^{jf}$ and $f^{jf}$ are weight matrix derived as follows:
\begin{gather}
	W^{jf}=(\bm{J}_{rd}^{js})^T\bm{J}_{rd}^{js} \\
	f^{jf}=(\dot{\bm{J}}_{rd}^{js}\dot{\bm{{\theta}}}^{js}-\ddot{p}_{rd})^T\bm{J}_{rd}^{js}
\end{gather}

\section{Whole Body Optimization}
\label{Sec Whole Body Optimization}
In the final stage of the optimization framework in this paper, an optimal control problem solver is employed to generate the whole-body trajectory, referencing the pre-generated trajectories from Sec. \ref{Sec SRMP Model Optimization} and Sec. \ref{Sec Joint Space Mapping Optimization}. As mentioned in Sec. \ref{Sec Symmetrical-Body Dynamic Model}, a 5-link dynamic system includes the trunk, thigh, calf, foot and arm is adopted.

\begin{figure}
	\centering
	\includegraphics[scale=0.37]{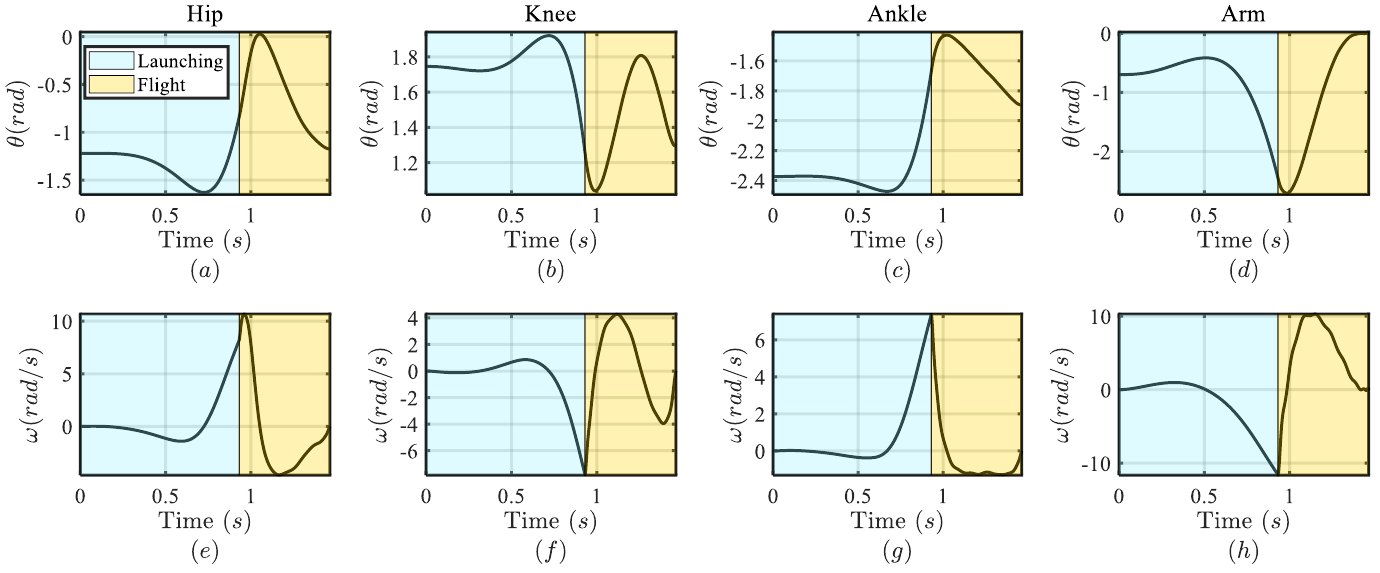}
	\caption{The optimized joint trajectories of Sec. \ref{Sec Whole Body Optimization}. Each column illustrates the joint angle $\theta$ and velocity $\dot{\theta}$ of one joint.}
	\label{Fig WB Joint Result}
\end{figure}
\begin{figure}
	\centering
	\includegraphics[scale=0.35]{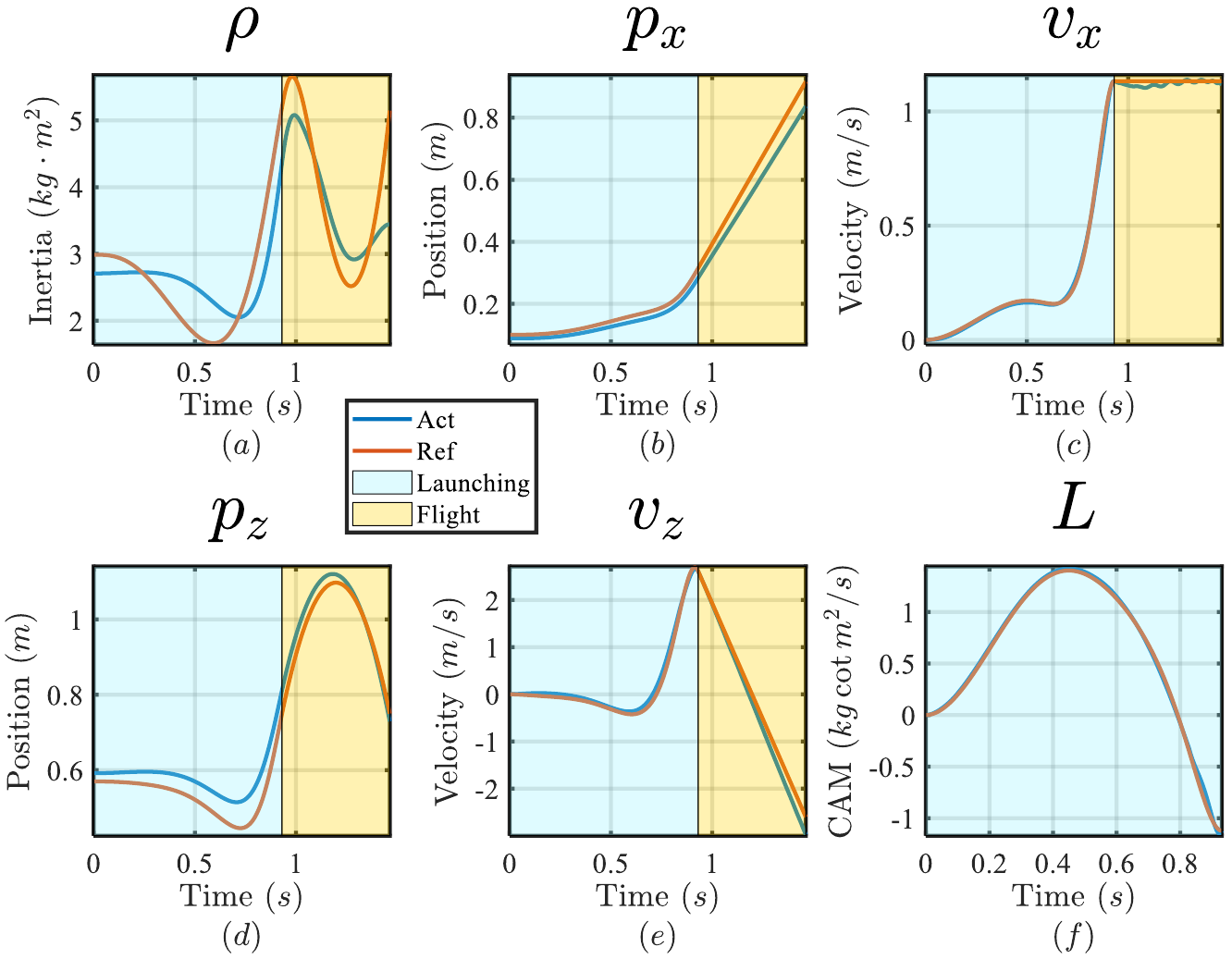}
	\caption{The optimized inertia and momentum trajectories of Sec. \ref{Sec Whole Body Optimization}.}
	\label{Fig WB Inertia and Momentum Result}
\end{figure}

The optimization in this section is also bifurcated into two parts, corresponding to the launching phase and the flight phase. Despite differences in the solver formulation between the two phases, the trajectories generated by Sec. \ref{Sec Joint Space Mapping Optimization} are uniformly input into both solvers. Results of this section are illustrated by curve graphs in Fig. \ref{Fig WB Joint Result} and Fig. \ref{Fig WB Inertia and Momentum Result}, and a snapshot in Fig. \ref{Fig WB Snapshot}.

\subsection{Launching Phase}
\subsubsection{Problem Formulation}
In this optimization, the state parameters $\bm{s}^{wl}$ consist of joint state, including joint angle, velocity and acceleration, which is listed as follows:
\begin{equation}
	\bm{s}^{wl}=[\bm{\theta}^{wb};\dot{\bm{\theta}}^{wb};\ddot{\bm{\theta}}^{wb}]
\end{equation}
Notably, as the configuration parameters like $\bm{\theta}^{wb}$ are utilized in both the launching and flight phases, no external distinctions on such parameters are adopted. Mixtures of superscripts $wl$, $wf$, and $wb$ are introduced to indicate that the same parameters (such as $\bm{\theta}^{wb}$) are employed in both the launching and flight phases.

\begin{figure}
	\centering
	\includegraphics[width=70mm,height=60mm]{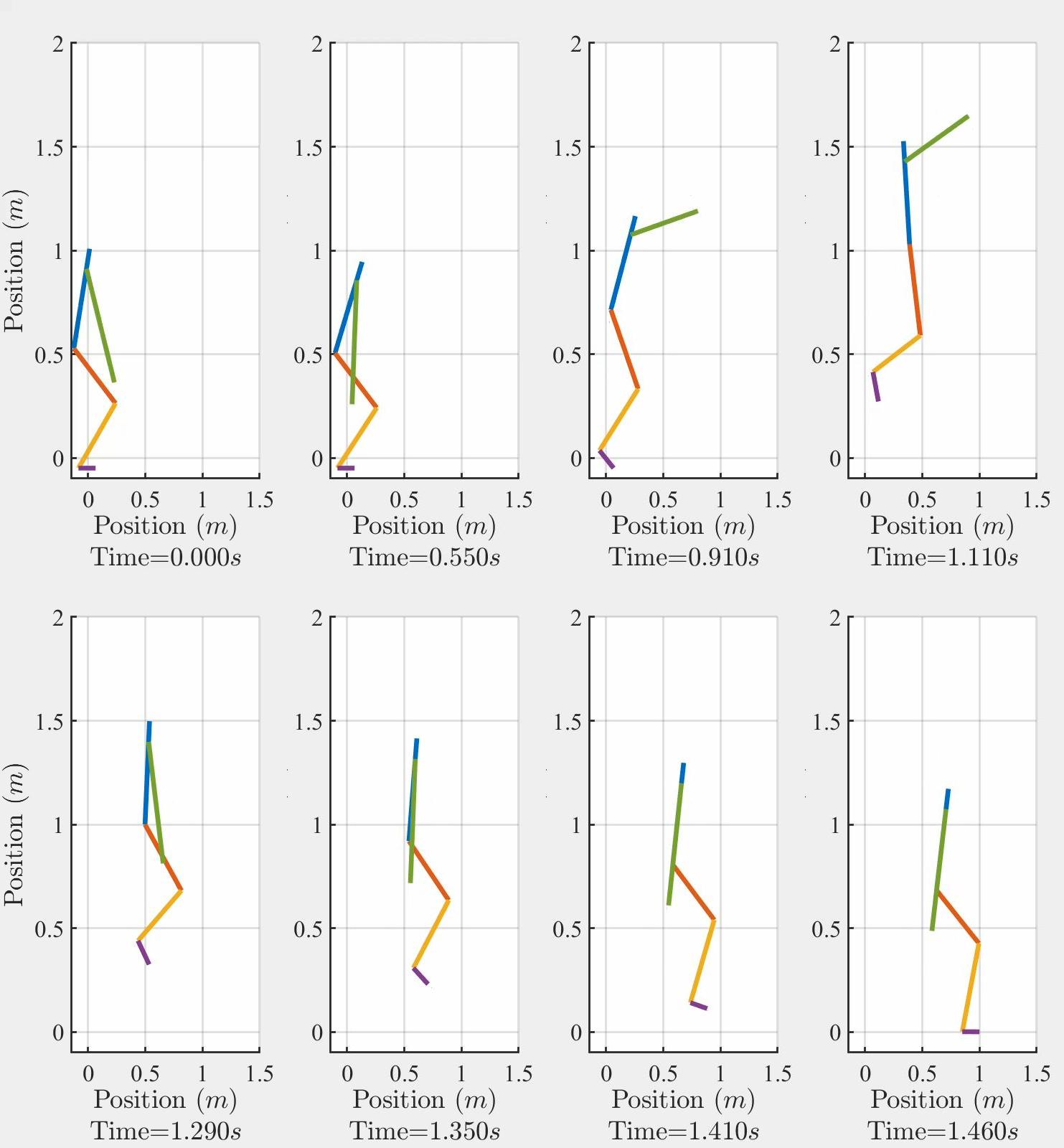}
	\caption{Illustration of the optimized results of Sec. \ref{Sec Whole Body Optimization}}
	\label{Fig WB Snapshot}
\end{figure}

In addition to the joint jerk, the control parameters $\bm{u}$ encompass the ground reaction force. Due to the dynamics error between the simplified model and the whole-body model, the solver might need to adjust the momentum while tracking trajectories to satisfy the constraints. Therefore, the ground reaction force is included in the control parameters to allow for potential adjustments by the solver. Consequently, the control parameters $\bm{u}^{wl}$ are outlined as follows:
\begin{equation}
	\label{Eq Control Parameter of WB Launching}
	\bm{u}^{wl}=[\dddot{\bm{\theta}}^{wb};F_x^{wb};F_z^{wb}]
\end{equation}
where $F_x^{wl}$ and $F_z^{wl}$ denote the ground reaction force in the $x$-axis and $z$-axis of this section, respectively. And boundary constraints are deployed for $\bm{s}^wl$ and $\bm{u}^{wl}$ as well.

\subsubsection{Cost Function}
\label{Sec WB Launching Cost Function}
The items of the cost function can be categorized into three groups: \emph{tracking constraints}, \emph{trajectory smoothness} and \emph{stability constraints}.

As implied by the term \textquotedblleft tracking constraints\textquotedblright, the initial segment of the cost function is formulated to improve the trajectory tracking performance of the solver. In this optimization problem, the solver is tasked with tracking the linear momentum and inertia trajectories. The dynamics constraints ensure the tracking performance of the angular momentum trajectory, with further details to be disclosed later. In summary, the primary focus in the cost function is on minimizing tracking errors:
\begin{align}
	\sum_{k=0}^{N_t^{wl}}(\|e_{\rho}[k]\|^2_{W_\rho^{wl}}+\|e_{\dot{\rho}}[k]\|^2_{W_{\dot{\rho}}^{wl}}+\|e_{v}[k]\|^2_{W_{v}^{wl}})
\end{align}
where $e_{\rho}$, $e_{\dot{\rho}}$ and $e_{v^{wl}}$ are the corresponding tracking error of $\rho^{ref}_{total}$, $\dot{\rho}^{ref}_{total}$ and $v^{ref}$, respectively. Taking $e_{\rho}$ for example, it is derived by $e_{\rho}=\rho^{ref}_{total}-\rho_{total}$. The CoM trajectories are anticipated to align with those generated by Sec. \ref{Sec SRMP Model Optimization}; hence, we use the same symbols for convenience in writing.

The smoothness of the output is guaranteed by the constraints on the state parameters and control parameters in the cost function. The joint acceleration, joint jerk and ground reaction force are minimized with designed weight, and the corresponding cost function can be illustrated as follows:
\begin{equation}
	\sum_{k=0}^{N_t^{wl}}(\|\ddot{\bm{\theta}}^{wl}[k]\|^2_{W_{\ddot{\theta}}^{wb}}+\|\bm{u}^{wl}[k]\|^2_{W_u^{wl}})
\end{equation}

Lastly, the heel's velocity is minimized to counteract possible exaggerated foot tilting:
\begin{equation}
	\sum_{k=0}^{N_t^{wl}}\|v_{heel\_z}^{wl}[k]\|^2_{W_{\ddot{h}}^{wl}}
\end{equation}
where $v_{heel\_z}^{wl}$ is calculated by the heel's Jacobian: $v_{heel}^{wl}=\bm{J}^{wb}_{heel}\dot{\bm{\theta}}^{wb}$

\subsubsection{Dynamics Constraints}
As mentioned in Eq. \ref{Eq Control Parameter of WB Launching}, the ground reaction force is defined as control parameters, hence the ground reaction torque $\bm{\tau}^{wb}_y$ is derived:
\begin{equation}
	\bm{\tau}^{wb}_y=\dot{L}^{ref}-(F_x^{wb}\ell_{rc\_z}-F_z^{wb}\ell_{rc\_x})
\end{equation}
It is worth noting that the CAM trajectory is considered by incorporating $\dot{L}_{ref}$ into this equation. This enables the solver to precisely track the CAM trajectory without the need for additional constraints or controllers. Subsequently, the formulation of the ground reaction force/torque vector $\bm{F}^{wl}_{ext}$ is as follows:
\begin{equation}
	\bm{F}^{wb}_{ext}=[F^{wb}_x;0;F^{wb}_z;0;\bm{\tau}^{wb}_y;0]
\end{equation}
and the floating base dynamics is constructed:
\begin{equation}
	\tilde{\bm{M}}^{wb}\ddot{\bm{\theta}}^{wb}+\tilde{\bm{H}}^{wb}=(\tilde{\bm{J}}^{wb})^T\bm{F}^{wb}_{ext}
\end{equation}

\subsubsection{Kinematic and Endpoint Constraints}
The kinematic and endpoint constraints are relatively straightforward to comprehend. The tiptoe is assumed to be fixed to simulate contact with the ground, and the heel is constrained to prevent it from sinking beneath the ground. Therefore, the kinematic constraints are formulated as follows:
\begin{equation}
	\left\{
	\begin{array}{l}
		\bm{J}^{wb}\dot{\bm{\theta}}^{wb}=0            \\
		\bm{J}^{wb}_{heel}\dot{\bm{\theta}}^{wb}\geq 0 \\
	\end{array}
	\right.
\end{equation}
Meanwhile, achieving the jump target requires precise control over the launching velocity of the CoM. Therefore, constraints are imposed on the CoM's velocity at the endpoint:
\begin{equation}
	v_{x\_end}^{wl}=v_{x\_end}^{wl\_ref},\quad v_{z\_end}^{wl}=v_{z\_end}^{wl\_ref}
\end{equation}
where $v_x^{wl}$ and $v_z^{wl}$ denote the velocity of the single-legged model's CoM in the $x$-axis and $z$-axis, respectively.

\begin{figure*}
	\centering
	\includegraphics[scale=0.12]{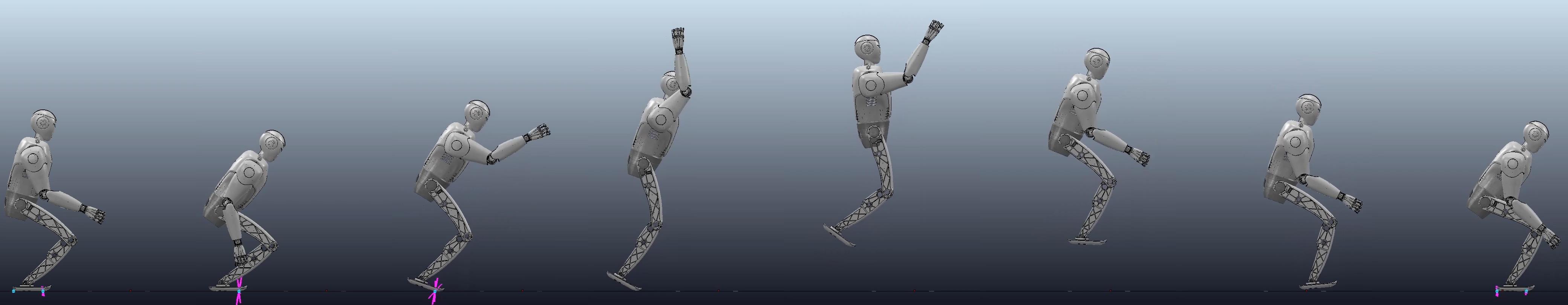}
	\caption{Snapshot of a forward jump simulation.}
	\label{Fig Application Simulation}
\end{figure*}

\begin{figure}
	\centering
	\includegraphics[scale=0.31]{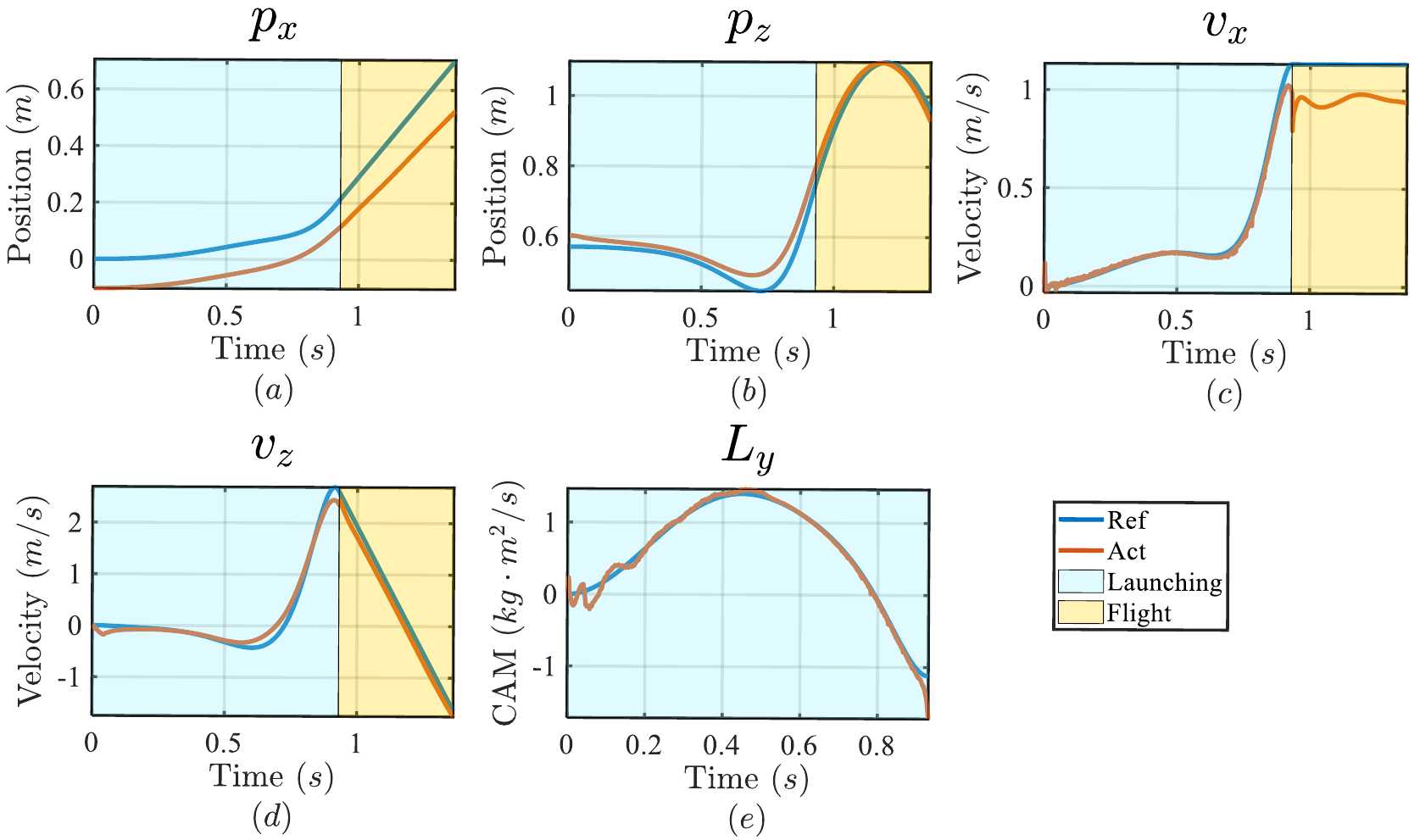}
	\caption{The momentum results of the simulation. Similar to previous figures, the launching and flight phases are distinguished by different colors.}
	\label{Fig Simulation Momentum Tracking}
\end{figure}

\subsection{Flight Phase}
\subsubsection{Problem Formulation}
The state parameters $\bm{s}^{wf}$ and control parameters $\bm{u}^{wf}$ are the same with those in the launching phase except for the ground reaction force no longer exists. Therefore, $\bm{s}^{wf}$ and $\bm{u}^{wf}$ can be listed as follows:
\begin{equation}
	\bm{s}^{wf}=[\bm{\theta}^{wb};\dot{\bm{\theta}}^{wb};\ddot{\bm{\theta}}^{wb}], \quad \bm{u}^{wf}=\dddot{\bm{\theta}}^{wb}
\end{equation}
Then, the boundary constraints are added as well.

\subsubsection{Cost Function}
The cost function of this phase is categorized into two groups: \emph{trajectory tracking} and \emph{trajectory smoothness}.

On one hand, the prioritized task of the solver for this phase is to track the reference inertia trajectory, enabling the robot to rotate as expected and attain an ideal landing posture. On the other hand, the tracking of momentum is no longer necessary due to the conservation of momentum. Consequently, the \emph{trajectory tracking} cost function can be listed as follows:
\begin{equation}
	\sum_{k=0}^{N_t^{wf}}(\|e_{\rho}[k]\|^2_{W_{\rho}^{wf}}+\|e_{\dot{\rho}}[k]\|^2_{W_{\dot{\rho}}^{wf}})
\end{equation}

\begin{figure*}
	\centering
	\includegraphics[scale=0.42]{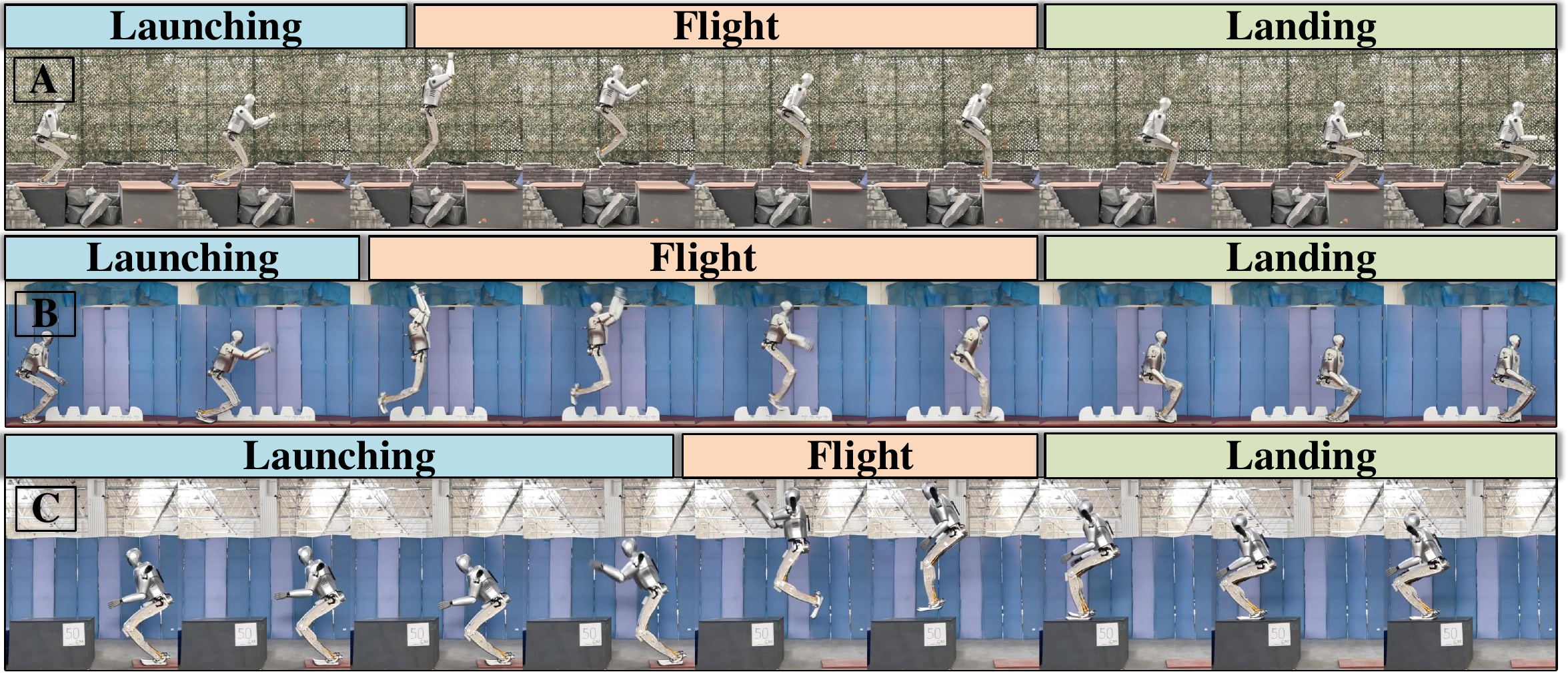}
	\caption{Snapshots of jump experiments, which are numbered by \textquotedblleft A, B, C\textquotedblright}
	\label{Fig Snapshot of forward jump experiment}
\end{figure*}

\begin{figure}
	\centering
	\includegraphics[scale=0.35]{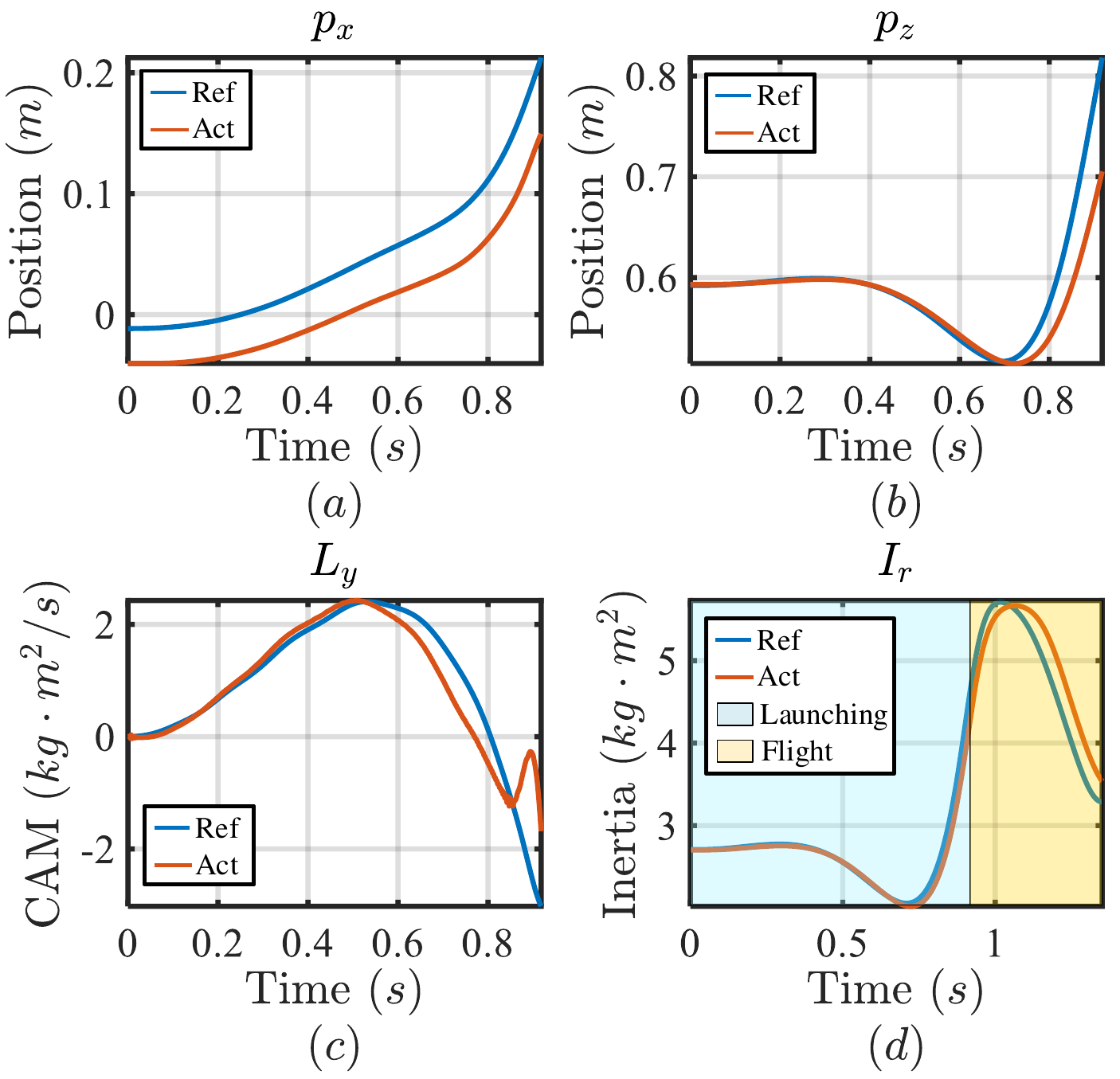}
	\caption{CoM position and CAM results during the launching phase, and inertia result during the launching and flight phases (experiment B).}
	\label{Fig Experiment Momentum}
\end{figure}

Similar to the cost function illustrated in Sec. \ref{Sec WB Launching Cost Function}, the smoothness of the output is guaranteed by minimizing the joint's acceleration and jerk. In addition, the ankle joint's velocity is minimized in this phase to prevent the robot from potential dramatic rotation of the foot. Therefore, the \emph{trajectory smoothness} part can be listed as follows:
\begin{equation}
	\sum_{k=0}^{N_t^{wf}}(\|\ddot{\bm{\theta}}^{wf}[k]\|^2_{W_{\ddot{\theta}}^{wf}}+\|\bm{u}^{wf}[k]\|^2_{W_{\bm{u}}^{wf}}+\|\dot{\theta}_{ankle}^{wf}[k]\|^2_{W_{ankle}^{wf}})
\end{equation}

\subsubsection{Linkage and Endpoint Constraints}
To link smoothly with the launching phase, the initial state of this phase is constrained to be equal to the final state of the launching phase as the linkage constraints:
\begin{equation}
	\left\{
	\begin{array}{l}
		\bm{\theta}^{wf}_{init}=\bm{\theta}^{wl}_{end}               \\
		\dot{\bm{\theta}}^{wf}_{init}=\dot{\bm{\theta}}^{wl}_{end}   \\
		\ddot{\bm{\theta}}^{wf}_{init}=\ddot{\bm{\theta}}^{wl}_{end} \\
	\end{array}
	\right.
\end{equation}
At the endpoint, the robot is supposed to be static and adjust to a proper posture to prepare for the impact. The static condition is achieved by constraining the state parameters at the endpoint, while the posture is constrained by the relative position between the foot and the CoM. At the endpoint, the robot is expected to swing its foot ahead its CoM to prepare enough room for the deceleration during the landing phase. By setting the relative position at the endpoint symmetrical with the one at the initial point, the solver is able to generate corresponding result.
\begin{equation}
	\left\{
	\begin{array}{l}
		\dot{\bm{q}}^{wf}_{end}=\ddot{\bm{q}}^{wf}_{end}=0                        \\
		p_{footx\_end}^{wb}-c_{x\_end}^{wb}=c_{x\_init}^{wb}-p_{footx\_init}^{wb} \\
	\end{array}
	\right.
\end{equation}

\section{Applications on Robot}
\label{Sec Applications on Robot}
\subsection{Introduction to the Platform}
\label{Sec Introduction to the Platform}
To demonstrate the proposed framework, we perform simulations with a humanoid robot platform Fig. \ref{Fig Application Robot Model}.

The robot is approximately 1.5 meters high and has a total mass of 42 kg and 14 degrees of freedom comprising 6 in each leg and 1 in each arm. A six-axis IMU sensor is mounted at the center of the trunk to measure the posture and acceleration of the robot body. The IMU has an embedded three-axis Micro-Electro-Mechanical System (MEMS) gyroscope with a full-scale range of 250$^\circ$/s and a three-axis MEMS accelerometer with a full-scale range of 8 g. Additionally, there are two six-dimensional force/torque sensors (M3714B2, Sunrise Instruments) mounted between the ankle and foot of each leg to obtain contact force/torque information. The force/torque sensor measures the three-dimensional force with and three-dimensional torque. In addition, the cycle time of the system is 1 ms.

\subsection{Simulation}
The simulation is conducted in the CoppeliaSim dynamic software\cite{coppeliaSim}, and the robot model is configured based on the platform (Sec. \ref{Sec Introduction to the Platform}). A forward jump is performed on the robot by applying the optimized whole-body trajectory to the corresponding joints in the position model, without employing any online controller or modification. The actuation parameters of the robot's joints are adjusted to ensure tracking performance in the position model. Additionally, the control framework used previously \cite{qi2023vertical} in the landing phase is employed in the simulation to obtain stable landing performance.

In the simulation, the robot successfully achieves the designed jump distance and height (Fig. \ref{Fig Application Simulation}). During the launching phase, the robot bends its upper body firstly to be ready for the acceleration process, which corresponds to the numeric illustration (Fig. \ref{Fig WB Snapshot}) and proves the practicality of the SRMP model. Simultaneously, it can be seen that the inertia shaping during the flight phase is illustrated by the retraction motion of the legs and arms, swinging the feet to desired landing point and obtains desired landing posture.

The momentum results of the simulation is illustrated in Fig. \ref{Fig Simulation Momentum Tracking}. In the figures of the CoM's position and velocity, differences between curves of the reference trajectories and the actual performance are evident. According to our analysis, such phenomenon dues to the differences of the dynamic models. As we mentioned in Sec. \ref{Sec Modeling of  the robot dynamics for the three-step optimization}, a 5-link model is used in the whole body optimization process. Although approximations have been made, we still cannot eliminate certain errors between it and the full-body dynamics model of the robot. Additionally, the dynamic parameters of the robot in the simulation are modified to have small differences with the dynamic models used in the optimization. With this modification, we try to simulate the inevitable error of the dynamic models between the real robot platform and the simulated one. Therefore, some errors are generated in the simulation. However, we find it to be acceptable because this problem can be solved by adding bias to the optimization's goal, and then the robot is able to achieve desired jump target in the experiment.

The angular momentum result (Fig. \ref{Fig Simulation Momentum Tracking}(e)) only illustrates the launching phase part, and the flight phase part is ignored due to its conservation. At the beginning of the launching phase, some vibration is captured, and then the curve of actual result smoothly coincides with the reference one. According to our analysis, this is cased by the non-ideal contact between the foot and the ground.

\subsection{Experiment}
We demonstrate the optimized trajectories on an adult-sized humanoid robot, which is the same platform introduced in our previous work\cite{qi2023vertical}. During the launching and flight phases, the robot runs under position model and executes the joint trajectories without any controller to demonstrate the performance of the optimized trajectory. When the robot meets the end of the trajectory, it switches to torque control model and waits for the landing impact. During the landing phase, the control framework introduced in \cite{qi2023vertical} is utilized to stabilize the robot.

Three of our forward jump experiments are illustrated in this paper (Fig. \ref{Fig Snapshot of forward jump experiment}). Experiment A and B both exceed the distance of 1.0 m, and in experiment C the robot jumps onto a 0.5 m high platform. In experiment A we set up a scene that simulates a wide ditch, while the scene of experiment B is designed to be clean for distance measurement.

Fig. \ref{Fig Experiment Momentum} illustrates the momentum and inertia data collected from Experiment B. The CoM and CAM data during the flight phase are not illustrated due to the lack of external sensors and state estimation algorithm. The reference data curve corresponds to the optimization results, while the actual data curve is derived from dynamics and kinematics analysis using sensor data collected from the robot. Visible errors are present due to modeling inaccuracies, sensor noise, and actuation errors. Notably, the robot does not reach the specified CoM height, and discrepancies are observed between the actual and reference CAM curves. Despite these errors, the robot successfully achieves the desired jumping target, suggesting that these discrepancies might be attributed to measurement errors. We believe that utilizing external sensors to measure the robot's kinematic information could yield more accurate results. Additionally, the inertia shaping trajectory effectively guides the robot to rotate ideally during the flight phase and land with proper foot placement, demonstrating the practicality of the framework's strategy.

\section{Conclusion}
\label{Sec Conclusion}
In this paper, we introduce a trajectory optimization framework that generates jump trajectory for a humanoid robot. The framework includes three parts and each part consists of an optimization algorithm and a dynamic model of the robot. In this framework, a SRMP model is introduced to balance the body posture and CAM during the launching phase optimization. Meanwhile, inertia shaping during the flight phase is considered and optimized to obtain ideal landing posture and foot placement. The framework is demonstrated by simulation and experiment on an adult-sized humanoid robot. However, we have been focused on single jump motions so far, and continuous jump is still unexplored. In our future work, we will dedicate to achieve continuous jump on our robot platform and make it more agile.

\bibliographystyle{IEEEtran} 
\bibliography{Reference} 

\begin{IEEEbiography}[{\includegraphics[width=1in,height=1.25in]{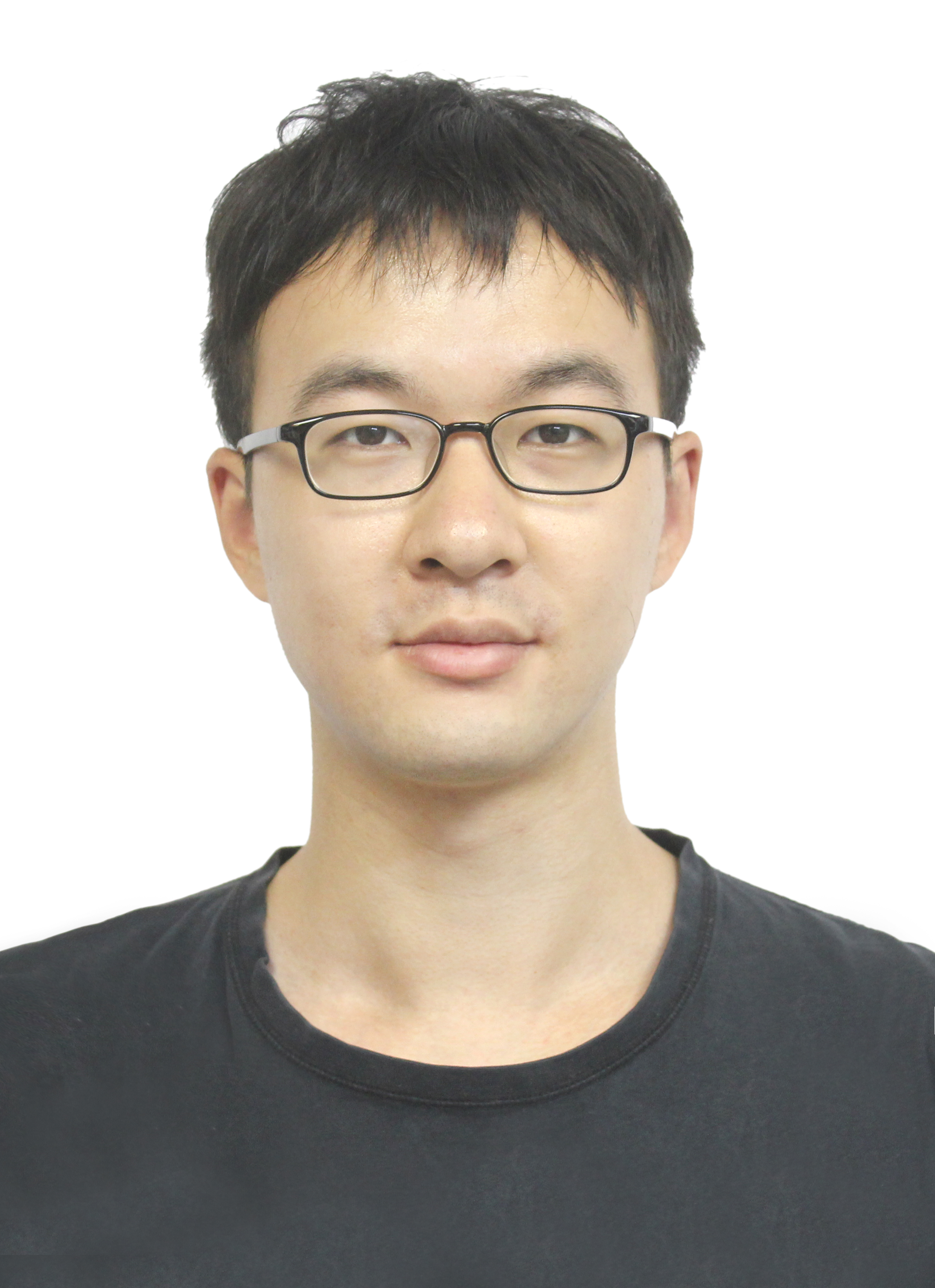}}]{Haoxiang~Qi}
	received B.S. degree in mechatronics engineering from the Beijing Institute of Technology (BIT), Beijing, China, in 2019.

	He is currently a Ph.D student with the Intelligent Robotics Institute, School of Mechatronical Engineering, Beijing Institute of Technology (BIT), Beijing, China.
\end{IEEEbiography}
\vspace{-12 mm}
\begin{IEEEbiography}[{\includegraphics[width=1in,height=1.15in]{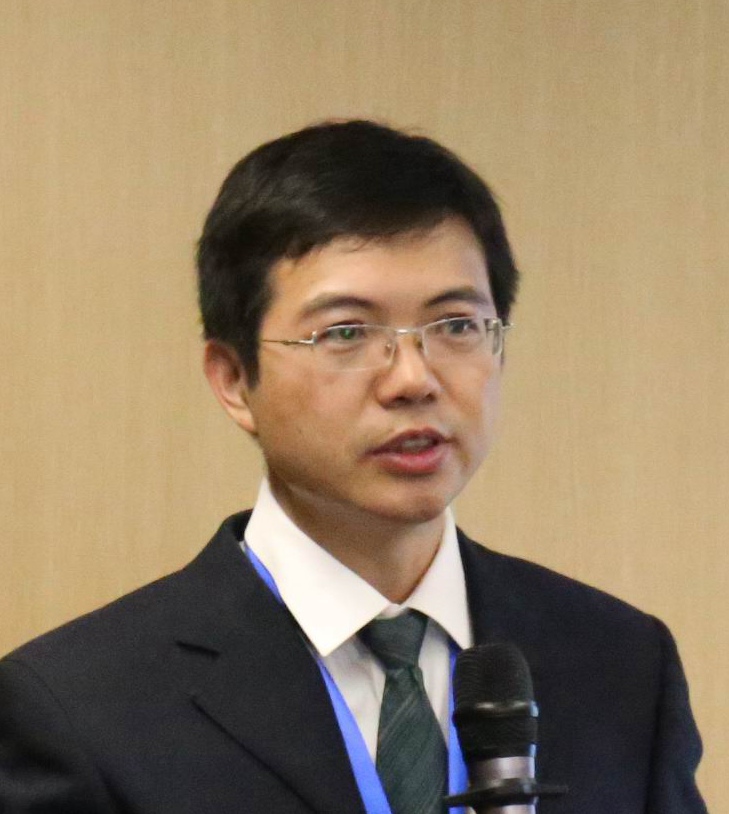}}]{Zhangguo~Yu}
	(Member, IEEE) received the B.S. degree in electronics engineering and the M.S. degree	in control engineering from the Southwest University of Science and Technology (SWUST), Mianyang, China, in 1997 and 2005, respectively, and the Ph.D. degree in mechatronics engineering from the Beijing Institute of Technology (BIT), Beijing, China, in 2009.

	He is currently a Professor with BIT, where he is also the Director of Intelligent Robotics Institute. His research interests include motion planning and control of biped robots.
\end{IEEEbiography}
\vspace{-12 mm}
\begin{IEEEbiography}[{\includegraphics[width=1in,height=1.25in]{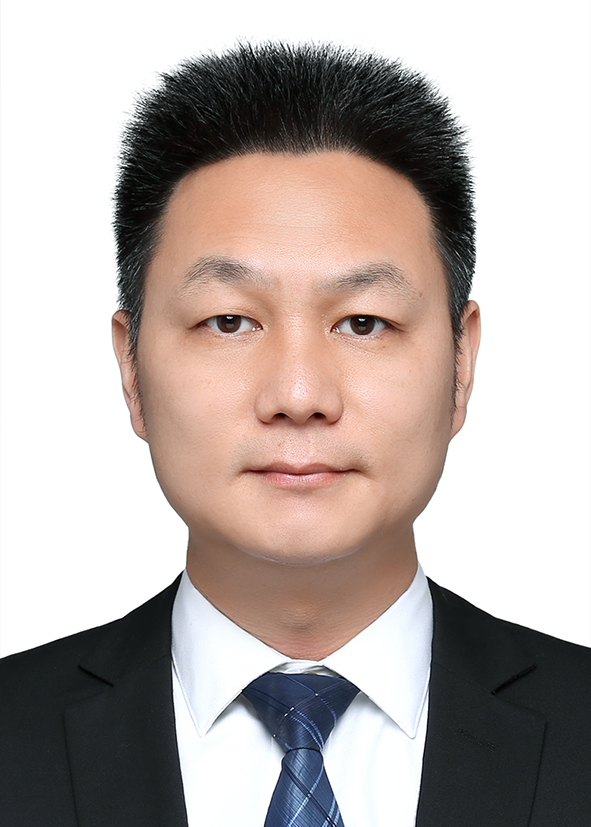}}]{Xuechao~Chen}
	(Member IEEE) received the B.S. and Ph.D. degrees in mechatronics engineering from Beijing Institute of Technology (BIT), China, in 2007 and 2013, respectively.

	He was a Visiting Student at the Robotics Institute, Carnegie Mellon University, Pittsburgh, USA, in 2012 and a Visiting Scientist at Sibley School of Mechanical and Aerospace Engineering, Cornell University, Ithaca, USA, in 2018. He was a Lecturer from 2013 to 2018, an Associate Professor from 2018 to 2021, and is currently a Professor at the School of Mechatronics Engineering, BIT.

	His research interests include biped locomotion, humanoid robot.
\end{IEEEbiography}
\vspace{-15 mm}
\begin{IEEEbiography}[{\includegraphics[width=1in,height=1.25in]{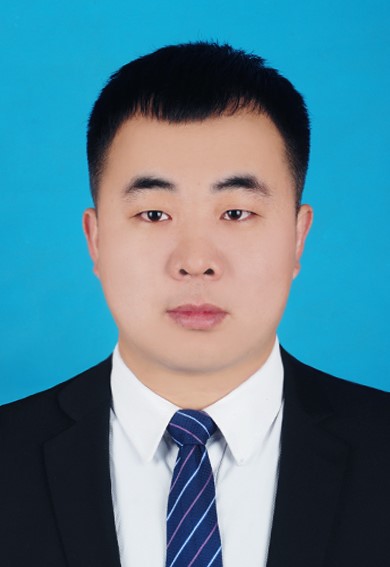}}]{Yaliang~Liu} 
	received the B.S. and M.S.degree in  mechatronics engineering in 2019 from the Yanshan University, Qinhuangdao, China. He is currently working toward a Ph.D. degree in mechatronics engineering from the Beijing Institute of Technology (BIT), Beijing, China.

	His research interests include actuator torque control and whole-body motion planning of humanoid robot.	
\end{IEEEbiography}
\vspace{-15 mm}
\begin{IEEEbiography}[{\includegraphics[width=1in,height=1.25in]{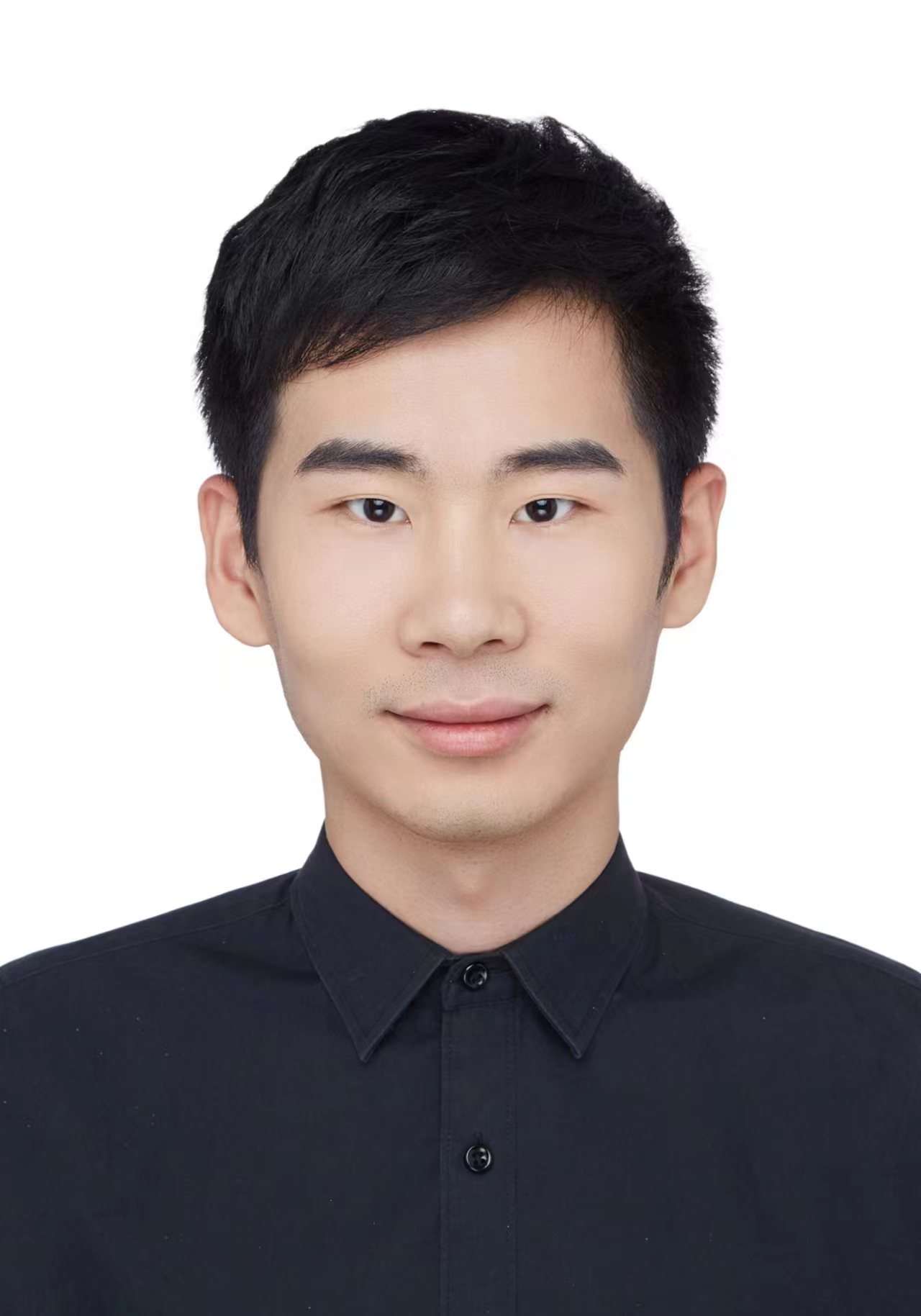}}]{Chuanku~yi} 
	Chuanku Yi received the B.S. degree in mechatronics engineering and the M.S. degree in mechanical engineering from the Beijing Institute of Technology, Beijing, China, in 2021 and 2024, respectively.  He currently works as an engineer at Qiyuan Lab, Beijing, China.	
\end{IEEEbiography}
\vspace{-15 mm}
\begin{IEEEbiography}[{\includegraphics[width=1in,height=1.25in]{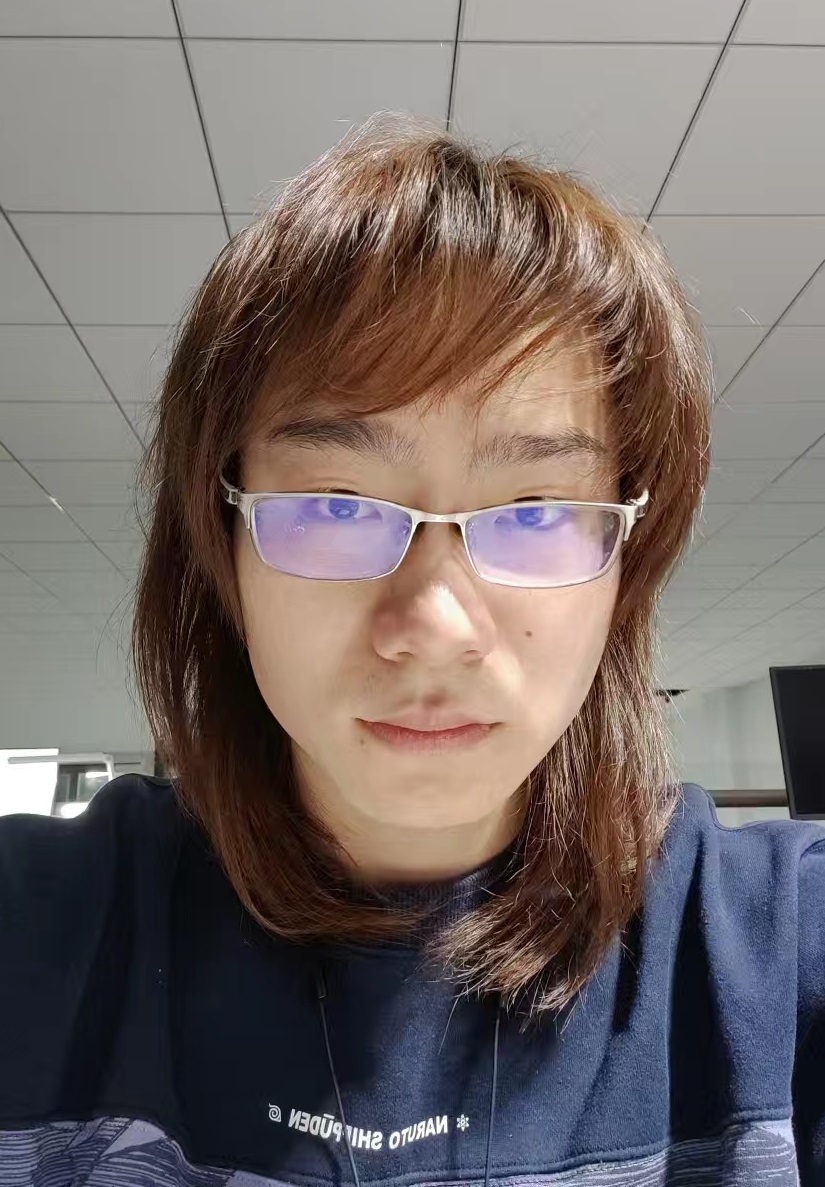}}]{Chencheng~Dong} 
	received the B.S. degree in mechatronics engineering from Beijing Institute of Technology (BIT), Beijing, China, in 2019. He is currently a Ph.D. candidate with the Intelligent Robotics Institute, School of Mechatronic Engineering, BIT, Beijing, China. His research interests include biped walking and running stability control, and compliance control.
\end{IEEEbiography}
\vspace{-15 mm}
\begin{IEEEbiography}[{\includegraphics[width=1in,height=1.25in]{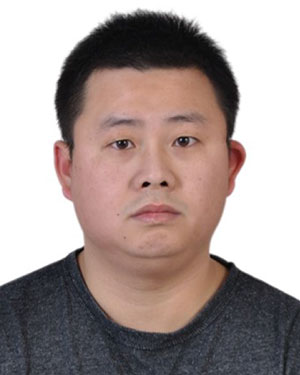}}] {Fei Meng} 
	received the B.S. and Ph.D. degrees in Mechatronics Engineering from Beijing Institute of Technology (BIT), Beijing, China, in 2008 and 2016, respectively. He was a Visiting Scholar in the Department of Mechanical Engineering and Intelligent Systems, University of Electro-Communications, Japan. He is currently an Assistant Professor at the School of Mechatronics Engineering, BIT. His research interest is motion control and actuator design for bionic robots.                                                                                    
\end{IEEEbiography}
\vspace{-15 mm}
\begin{IEEEbiography}[{\includegraphics[width=1in,height=1.25in]{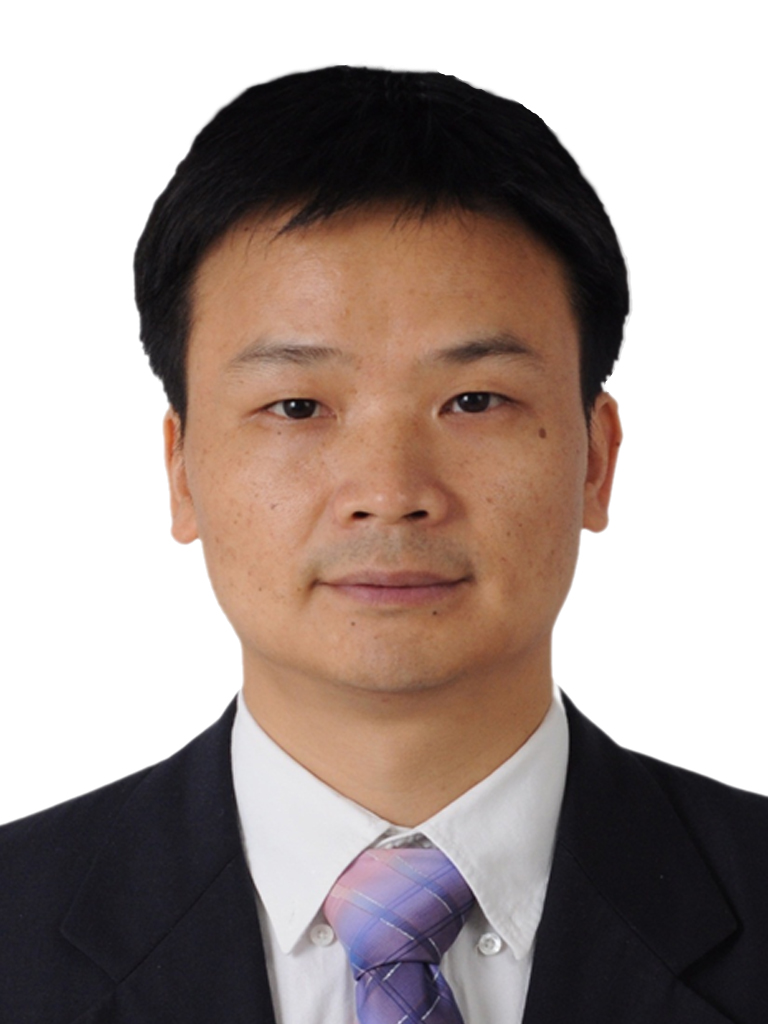}}]{Qiang~Huang}
	(Fellow, IEEE) received the B.S. and M.S. degrees in electrical engineering from the Harbin Institute of Technology, Harbin, China, in
	1986 and 1989, respectively, and the Ph.D. degree in mechanical engineering from Waseda University, Tokyo, Japan, in 1996. 
	
	From 1996 to 1999, he was a Research Fellow with the National Institute of Advanced Industrial Science and Technology, Japan. From 1999 to 2000, he was a Research Fellow with the University of Tokyo, Japan.	Since 2000, he has been a Professor with the Beijing Institute of Technology, Beijing, China, where he is currently the Executive Director of the Beijing Advanced Innovation Center for Intelligent Robots
	and Systems. His research interests include biped locomotion and biorobotic systems.

	Dr. Huang was the recipient of the IFToMM Award of Merit.
\end{IEEEbiography}

\end{document}